\documentclass[11pt, a4paper]{article}
\usepackage{hybridreport}
\usepackage{needspace}
\usepackage{placeins}
\usepackage{tikz}
\usetikzlibrary{arrows.meta,positioning,calc,fadings,shadows.blur}
\usepackage{pgfplots}
\pgfplotsset{compat=1.16}
\usepgfplotslibrary{groupplots}
\newcommand{\PowerLawA}{0.063}
\newcommand{\PowerLawAlpha}{0.035}
\newcommand{\PowerLawRsq}{0.98}
\newcommand{\PowerLawDoubling}{2.4}

\pgfplotsset{/pgf/number format/assume math mode=true}

\DeclareTextFontCommand{\textbf}{\fontsize{10.5}{13}\selectfont\bfseries\sffamily}

\newcommand{\pio}{$\pi_{0.5}$}
\newcommand{\projectpage}{\href{https://a-0.airoa.io}{\mbox{a-0.airoa.io}}}
\newcommand{\piz}{$\pi_{0}$}
\newcommand{\Lc}{\mathcal{L}}
\newcommand{\gatetile}[2]{\begin{tabular}[t]{@{}c@{}}%
  \includegraphics[width=0.193\linewidth]{figures/gate/#1.jpg}\\[-1.5pt]#2\end{tabular}}

\newfontfamily\airmark{Manrope-ExtraBold.ttf}[Path=fonts/]
\newcommand{\teaserbars}{tables/benchmark_bars}

\newcommand{\teaserbody}{\sffamily\fontsize{7.1}{9.3}\selectfont\hyphenpenalty=10000\exhyphenpenalty=10000}
\newcommand{\bmbar}[4]{%
  \pgfmathsetmacro{\bmy}{-#1*0.30}%
  \ifnum#4=1\relax
    \def\bmcol{Primary}\def\bmtxt{Primary}\def\bmwt{\bfseries}%
  \else
    \def\bmcol{AccentMuted}\def\bmtxt{TextBody}\def\bmwt{}%
  \fi
  \node[anchor=east, inner sep=0pt, text=\bmtxt, text height=1.5ex, text depth=0.5ex,
        font=\sffamily\bmwt\fontsize{7.1}{8.4}\selectfont] at (2.30,\bmy) {#2};
  \if\relax\detokenize{#3}\relax\else
    \pgfmathsetmacro{\bmw}{3.2*#3/100}%
    \fill[\bmcol] (2.45,\bmy+0.08) rectangle ++(\bmw,-0.16);
    \node[anchor=west, inner xsep=1.5pt, inner ysep=0.6pt, fill=white, text=\bmtxt,
          font=\sffamily\bmwt\fontsize{6.6}{8}\selectfont] at (2.45+\bmw+0.03,\bmy) {#3};
  \fi}

\newcommand{\bmhead}[1]{\mbox{\sffamily\bfseries\fontsize{8.4}{11}\selectfont #1}}

\newcommand{\bmavg}[1]{{\sffamily\bfseries\fontsize{8.4}{11}\selectfont #1}}
\newcommand{\bmnum}[1]{\makebox[1.7em][r]{#1}}

\newcommand{\bmbest}[1]{\makebox[1.7em][r]{\rmfamily\bfseries #1}}
\newcommand{\bmnone}{\makebox[1.7em][r]{\color{RuleGray}--\kern0.25em}}
\newcommand{\bmgap}{\leavevmode\kern4pt}
\newcommand{\bmdatahead}[1]{\mbox{\sffamily\fontsize{8.4}{11}\selectfont\color{TextMuted}#1}}
\newcommand{\bmdata}[1]{\leavevmode{\color{TextMuted}#1}}
\newcommand{\bmthinrule}{\arrayrulecolor{RuleGray}\specialrule{0.4pt}{0pt}{0pt}\arrayrulecolor{black}}

\reporttype{}
\logo{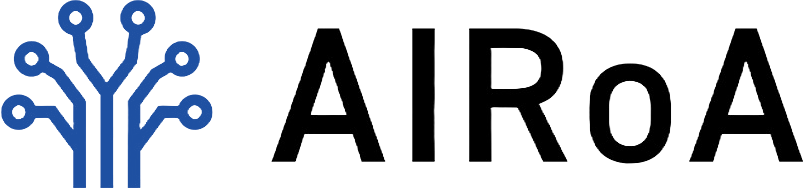}
\title{{\fontsize{30}{34}\selectfont AIRoA-0}\\[1pt]%
  {\fontsize{17}{22}\selectfont\mdseries Long-Horizon, Precise, Cross-Embodiment\\[-6pt]%
  Manipulation from Handheld Data}}
\shorttitle{AIRoA-0: Long-Horizon, Precise, Cross-Embodiment Manipulation from Handheld Data}
\hypersetup{pdftitle={AIRoA-0: Long-Horizon, Precise, Cross-Embodiment Manipulation from Handheld Data},
  pdfauthor={AIRoA}}
\author{\small\mdseries
  \renewcommand{\arraystretch}{1.3}\setlength{\tabcolsep}{4pt}%
  \begin{tabular}{@{}p{0.235\linewidth}p{0.235\linewidth}p{0.235\linewidth}p{0.235\linewidth}@{}}
    Yueh-Hua (Kris) Wu\textsuperscript{*\dag} & Kei Ota\textsuperscript{*} & Hanlong Li & Hayata Yuasa \\
    Jason Wang & Kengo Ikeuchi & Kensuke Chiyoma & Makoto Sugiura \\
    Masafumi Takahashi & Masatoshi Tateno & Naoaki Kanazawa & Nicola Scianca \\
    Petr Khrapchenkov & Satoshi Otsubo & Takehiko Ohkawa & Takuya Okubo \\
    Tatsuya Matsushima & Yuki Noguchi & Yuma Shin & Zixuan Zhong \\[3pt]
  \end{tabular}%
}
\institution{\sffamily AIRoA\quad \textsuperscript{*}Project Lead\quad \textsuperscript{\dag}Research Lead}
\date{November 2026}

\abstracttext{%
  \mbox{AIRoA-0} (\mbox{A-0}) is a bimanual manipulation system learned from 16,151 hours of
  handheld demonstrations, with no robot data collected at any stage. The
  demonstrations are recorded with YUBI, our open-source handheld gripper,
  which lets people record manipulation data anywhere with no robot present.
  From a single instruction, \mbox{A-0} sorts more than a dozen small parts into
  their numbered cells and changes the mop pads and brush of a robot vacuum;
  among other tasks, it unzips a pencil case and folds a towel. It performs
  these tasks using only a camera on each wrist, with no head or scene camera,
  at a site where none of its training data was recorded. Across the ten tasks
  of our evaluation, it averages a success rate of 78\%, against 27\% for the
  strongest of four other vision-language-action models post-trained on the
  same task demonstrations. We collect no
  teleoperation data and use no reinforcement learning or on-robot corrections, and the robot
  appears only at test time. The system learns which regions of each wrist
  view matter for the task, predicts the stage the task has reached, and
  decodes actions as a continuous stream. \mbox{A-0} is embodiment-agnostic: it
  runs on robots that mount the YUBI gripper, such as a bimanual Franka, TX-G2,
  OpenArm and UR. We are releasing the dataset, to our knowledge the largest open
  dataset of handheld-gripper demonstrations with task and subtask
  instructions, together with the pre-trained weights and the post-training
  recipe. Videos of \mbox{A-0} on every task are on the project page,
  \projectpage. Anyone can build the gripper, record a task of their own and
  post-train the model on it in under four hours on a single node of eight H200 GPUs.
}

\begin{document}
\maketitle

\section{Introduction}

\suppressfloats[t]
\begin{figure}[t]
  \centering
  \input{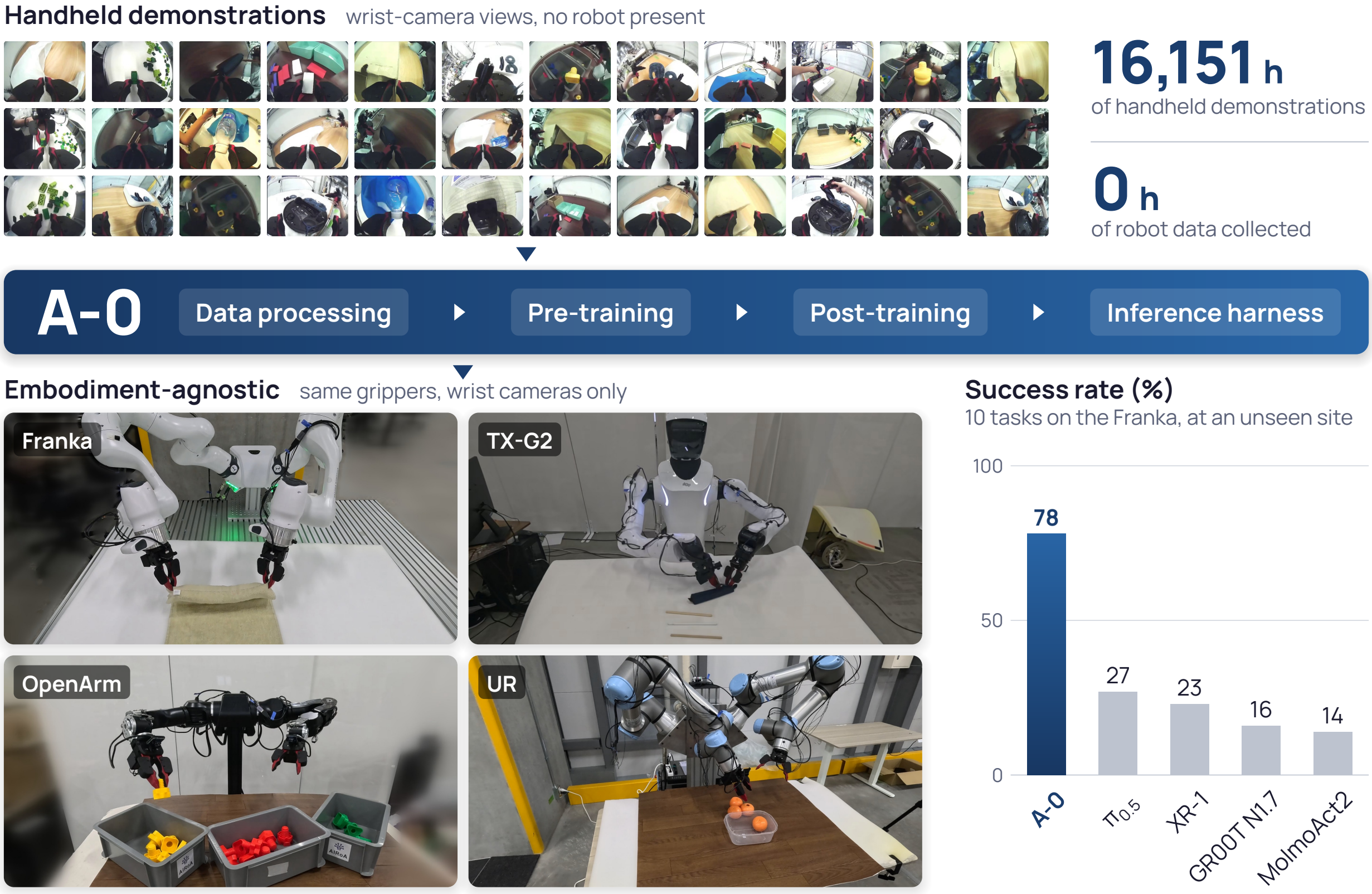}
  \caption{\mbox{A-0} in overview. Handheld demonstrations, seen here through their
  wrist cameras, are the only data we collect. The four parts of the system
  turn them into actions for robots that mount the same gripper and cameras;
  the photographs show towel folding on the Franka, the
  pencil case on TX-G2, colour sorting on OpenArm and the orange pyramid on UR. The chart gives each model's success rate, averaged over the ten tasks
  of Section~\ref{sec:bench}.}
  \label{fig:teaser}
\end{figure}

Data recorded with handheld grippers grows with the number of people who
record it, not with the number of robots available: a person holds a gripper
in each hand and does the task, and no robot is needed. UMI~\cite{umi}
introduced the approach; with YUBI~\cite{yubi}, our open-source gripper of
this kind, we have recorded 16,151 hours of bimanual demonstrations, 10.65
million subtask segments in all, each with a task instruction and a subtask
instruction. \mbox{AIRoA-0} (\mbox{A-0}) uses handheld demonstrations for pre-training and
post-training alike: no teleoperation, no reinforcement learning, no on-robot
corrections and no simulation.

The properties that let this data scale are the ones that make it hard to
learn from. Every demonstration is made by a person, so nothing in the data
shows a robot acting, failing or recovering. The policy sees only through the
cameras on the grippers, the one view that is the same in a person's hand and
on a robot: with no head or scene camera, it never sees the workspace as a
whole and must follow a long task through two partial views that are often
consistent with more than one of its steps. Data that anyone can record is
the work of many operators, here across well over a hundred tasks, and a
single model has to learn from all of it. And people record wherever they
are, which is not where the robot will work. Deployment at an unseen site is
therefore the ordinary condition of this approach and the one under which it
must be judged: not one of the demonstrations \mbox{A-0} learns from was recorded
at the site where it is evaluated.

\mbox{A-0} works under all of these conditions at once. Given a single sentence of
instruction, it sorts more than a dozen small parts into their numbered
cells, deciding for each part which cell it belongs in, which hand should
take it, and whether it must first pass from one hand to the other. It opens
the zipper of a pencil case, which the two grippers must grasp almost side
by side; it folds a towel; it replaces the mop pads and the brush of a robot
vacuum. These are long tasks that
demand precise contact, and the robot performs them with the grippers that
people record with. On the ten tasks of our evaluation, \mbox{A-0} averages a
success rate of 78\%, against 27\% for the strongest of four other
vision-language-action models post-trained on the same task demonstrations:
each of them with its best setting on every task, \mbox{A-0} with one recipe
for all ten. And although nothing in its data shows a robot meeting a
disturbance, \mbox{A-0} acts on what it sees: when a person pushes the bins aside
as it brings a part to them, it moves over to the correct bin in one continuous
motion (Figure~\ref{fig:perturb}). Because those grippers mount on different arms, \mbox{A-0}
is embodiment-agnostic: the same weights run on robots such as a bimanual
Franka, TX-G2, OpenArm and UR (Section~\ref{sec:robots}).

\mbox{A-0} is built as one system that spans every layer between the gripper in a
person's hand and the motion of the robot (Figure~\ref{fig:pipeline}), and
each of these difficulties is met at the layer where it arises. State and
action are defined at the grippers themselves, so the same numbers describe a
person holding them and a robot carrying them, on one arm as on another.
Pre-training tells a single model, through language at two levels, both the
task and the step in progress, and a visual evidence budget~\cite{s2} teaches
it, with no annotation of where to look, which regions of each wrist view
matter for the task. Post-training adds a prediction of the stage a long task
has reached and a memory of recent motion, and the inference harness smooths
the predicted stage and decodes actions as a continuous stream on a
non-uniform noise schedule that we propose.

With this report we are releasing the full dataset of 16,151 hours of
handheld-gripper demonstrations with task and subtask instructions, together
with the pre-trained weights of \mbox{A-0} and its post-training recipe. Together with the YUBI gripper's hardware design
and collection software, which are already public, they open the path
from a gripper in a person's hand to a policy on a robot: the gripper is built
from the open design, the task is recorded by hand, and the pre-trained model
is post-trained on it with the one recipe that serves every task in this
report, in under four hours on a single node of eight H200 GPUs.

Our contributions are:

\begin{enumerate}
  \item \mbox{A-0}, a bimanual manipulation system learned from 16,151 hours of
    handheld data, with no robot data collected at any stage.
  \item An open release of the path from a handheld gripper to a robot
    policy: the dataset, to our knowledge the largest open dataset of
    handheld-gripper demonstrations with task and subtask instructions; the
    pre-trained weights; and the post-training recipe, alongside the public
    YUBI hardware design and collection software.
  \item Long, precise tasks carried out from a single instruction, through
    two wrist cameras, at a site where none of the training data was
    recorded.
  \item An embodiment-agnostic system: it runs on robots that take the
    recording gripper as their end effector, such as a bimanual Franka,
    TX-G2, OpenArm and UR.
\end{enumerate}

\section{Results}
\label{sec:results}

\subsection{Ten tasks}
\label{sec:bench}

\begin{figure}[p]
  \centering
  \input{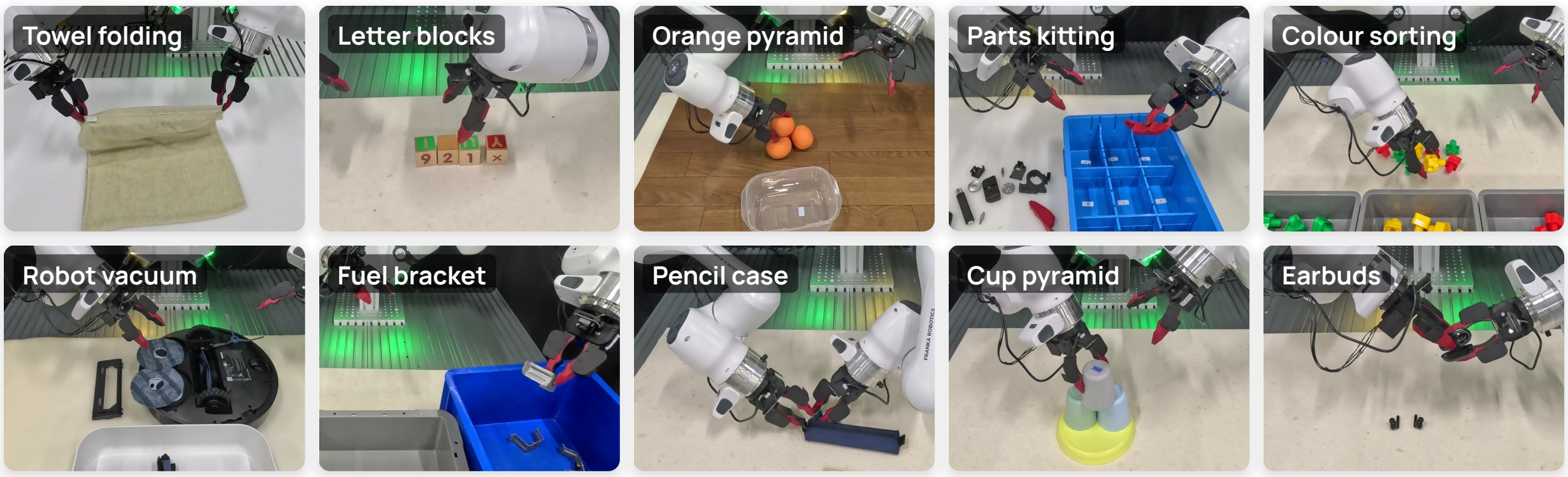}
  \caption{The ten evaluation tasks, each in a frame from an evaluation run of
  \mbox{A-0}.}
  \label{fig:tasks}
\end{figure}

On the ten bimanual tasks of Figure~\ref{fig:tasks}, evaluated on a Franka platform at
a site where none of the training data was recorded, \mbox{A-0} averages a
success rate of 78\%, against 14\% to 27\% for four other
vision-language-action models (Table~\ref{tab:benchmark}), and it has the
highest success rate on every task, from sorting parts and stacking cups to
folding a towel and servicing a robot vacuum. Each task is specified by a
single instruction, given once at the start. Each \keyword{success rate} is
an average over 15 trials, with partial credit for a task that is partly
completed. Post-training needs
little data for each task: six of the ten tasks use between three and six
hours of demonstrations, and only the cup pyramid and the earbuds use more than ten
(Table~\ref{tab:benchmark}).

The four models are \pio~\cite{pi05}, \mbox{Xiaomi-Robotics-1}
(XR-1)~\cite{xiaomi}, GR00T~N1.7~\cite{groot17,groot} and
MolmoAct2~\cite{molmoact2}, each
starting from its public pre-trained checkpoint and post-trained on the same
task demonstrations as \mbox{A-0}. Each of them is post-trained on every task
over a range of settings, with action horizons of 4, 6, 8, 10, 12, 14 and 16,
several action frequencies, and 60,000, 80,000 and 100,000 updates, and on
every task we report its best result. \mbox{A-0} is given no such search: every
task is post-trained with the one recipe of Table~\ref{tab:posttrain}.

\begin{table}[p]
  \centering
  \caption{The ten evaluation tasks: the hours of handheld demonstrations of
  each task that post-training uses, and the success rate (\%), each an
  average over 15 trials. \mbox{A-0} uses the one recipe of
  Table~\ref{tab:posttrain} on every task; the other four models use their best
  setting for each task. The best entry for each task is in bold.}
  \label{tab:benchmark}
  \vspace{0.5em}
  \small
  \setlength{\aboverulesep}{0pt}\setlength{\belowrulesep}{0pt}%
  \setlength{\tabcolsep}{0pt}\renewcommand{\arraystretch}{1.3}%
  \begin{tabular}{@{}p{0.17\textwidth}>{\centering\arraybackslash}p{0.10\textwidth}>{\columncolor{AccentLight}\centering\arraybackslash}p{0.1460\textwidth}*{4}{>{\centering\arraybackslash}p{0.1460\textwidth}}@{}}
\toprule
 & \bmdatahead{Data (h)} & \bmhead{A-0} & \bmhead{π\textsubscript{0.5}} & \bmhead{XR-1} & \bmhead{GR00T N1.7} & \bmhead{MolmoAct2} \\
\midrule
Towel folding & \bmdata{3.3} & \bmbest{73} & \bmnum{22} & \bmnum{21} & \bmnum{0} & \bmnum{8} \\
Letter blocks & \bmdata{3.4} & \bmbest{84} & \bmnum{13} & \bmnum{0} & \bmnum{24} & \bmnum{0} \\
Orange pyramid & \bmdata{3.4} & \bmbest{90} & \bmnum{2} & \bmnum{52} & \bmnum{13} & \bmnum{28} \\
Parts kitting & \bmdata{4.0} & \bmbest{83} & \bmnum{30} & \bmnum{13} & \bmnum{7} & \bmnum{12} \\
Colour sorting & \bmdata{4.1} & \bmbest{97} & \bmnum{45} & \bmnum{39} & \bmnum{20} & \bmnum{19} \\
Robot vacuum & \bmdata{5.3} & \bmbest{73} & \bmnum{1} & \bmnum{14} & \bmnum{1} & \bmnum{3} \\
Fuel bracket & \bmdata{6.3} & \bmbest{93} & \bmnum{34} & \bmnum{33} & \bmnum{47} & \bmnum{11} \\
Pencil case & \bmdata{8.8} & \bmbest{73} & \bmnum{36} & \bmnum{31} & \bmnum{15} & \bmnum{23} \\
Cup pyramid & \bmdata{17.9} & \bmbest{80} & \bmnum{71} & \bmnum{11} & \bmnum{23} & \bmnum{36} \\
Earbuds & \bmdata{25.8} & \bmbest{30} & \bmnum{13} & \bmnum{18} & \bmnum{10} & \bmnum{3} \\
\bmthinrule
\bmavg{Average} & & \bmbest{78} & \bmnum{27} & \bmnum{23} & \bmnum{16} & \bmnum{14} \\
\bottomrule
\end{tabular}

\end{table}

\begin{figure}[p]
  \centering
  \input{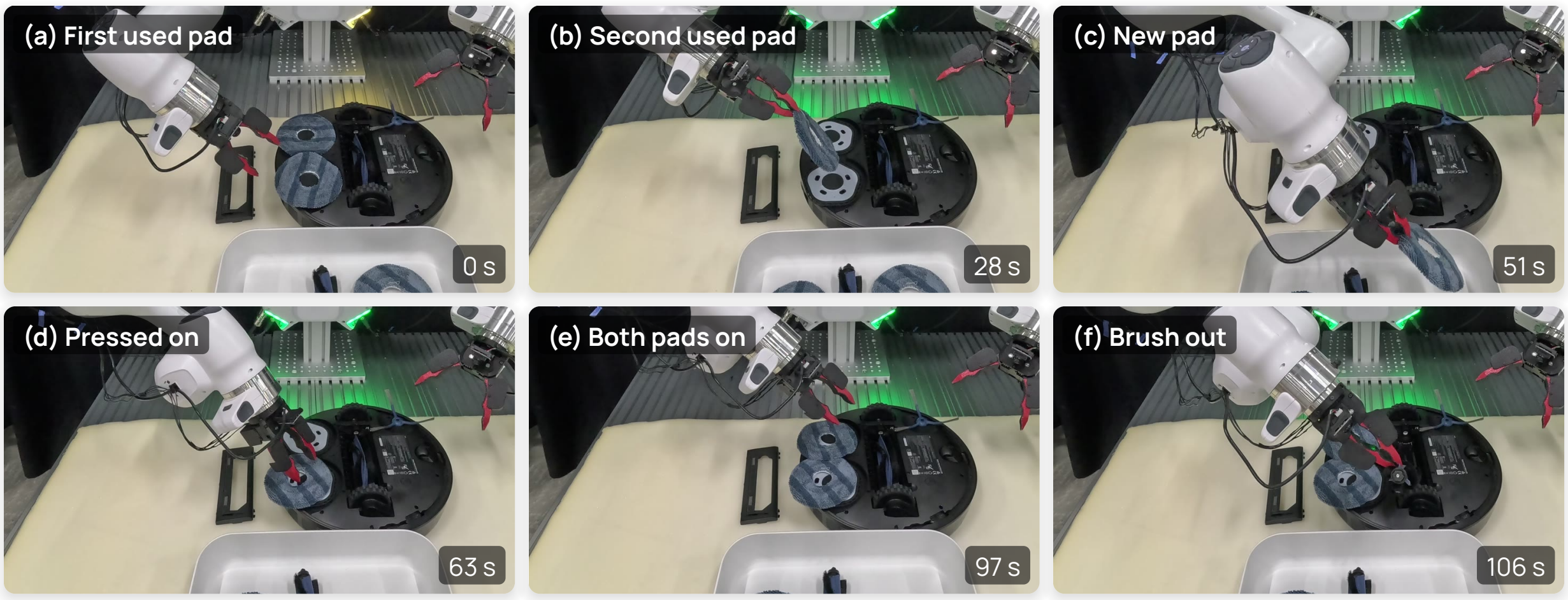}
  \caption{Two minutes of one autonomous run of the robot-vacuum task, with the
  time on the robot since (a). (a) The gripper takes hold of the first used
  mop pad, and (b) the second comes off. (c) A new pad is carried from the box
  and (d) pressed onto its mount; (e) both new pads are on, and (f) the brush
  is pulled out.}
  \label{fig:longtask}
\end{figure}

\Needspace{5\baselineskip}
Several of the tasks run for minutes: in the two minutes of
Figure~\ref{fig:longtask}, \mbox{A-0} takes the two used mop pads off a robot
vacuum, attaches two new ones from the box and pulls out the brush, one step
after another, from the single instruction it was given at the start. Videos
of every task are on the project page.

\Needspace{8\baselineskip}
\subsection{A changing scene}
\label{sec:perturb}

\begin{figure}[H]
  \centering
  \input{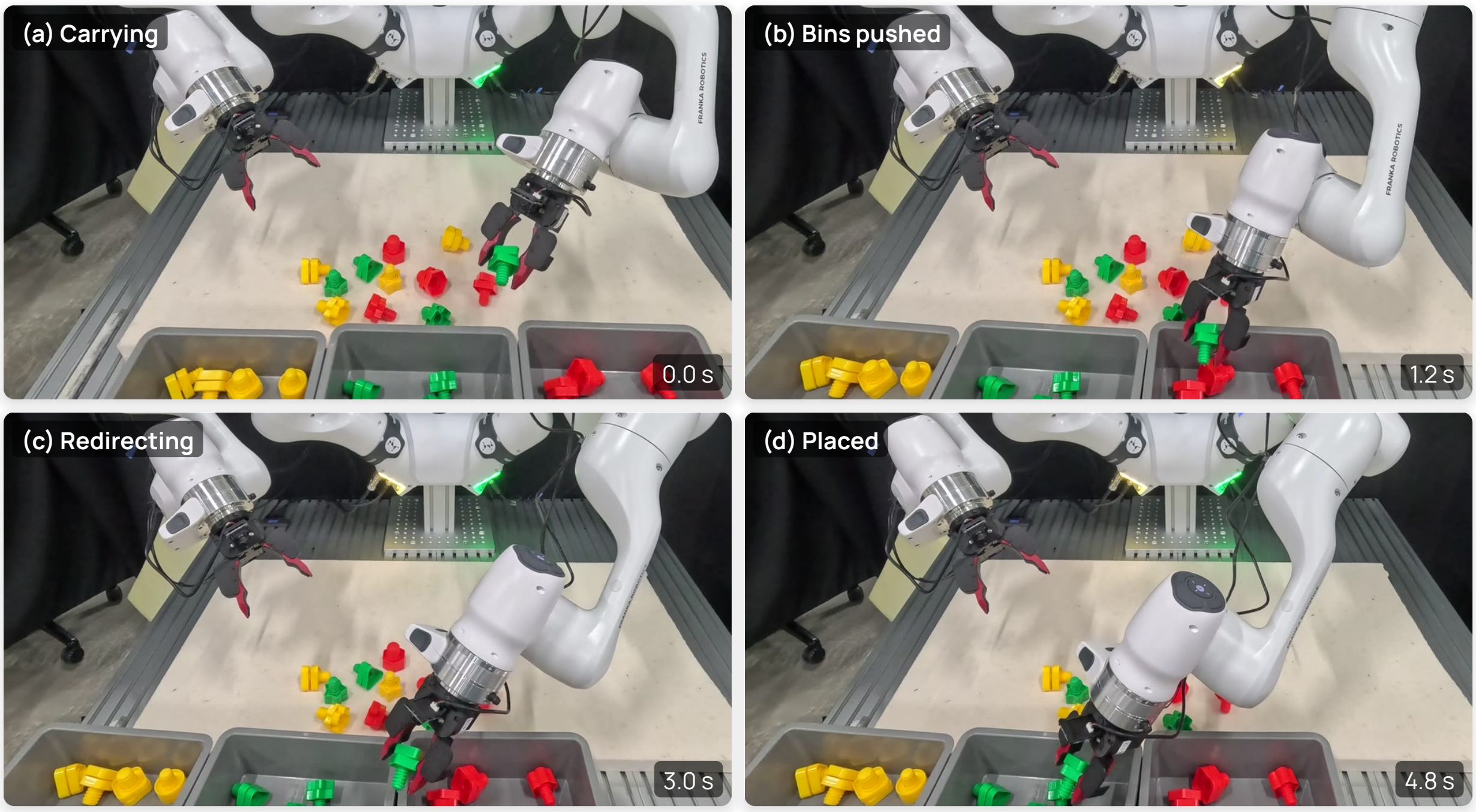}
  \caption{The bins moved during autonomous colour sorting, with the time on
  the robot since (a). (a) An arm
  carries a green part to the green bin; the bins are, from left to right,
  yellow, green and red. (b) A person pushes the bins to the left, and the
  red bin is now under the gripper. (c) The arm moves over to the green bin
  and (d) drops the part into it.}
  \label{fig:perturb}
\end{figure}

Nothing in the training data shows a robot meeting a disturbance, and a
policy that replayed its demonstrations would carry on with the plan it had
made. \mbox{A-0} acts on what its wrist cameras see. In
Figure~\ref{fig:perturb}, a person pushes the row of bins aside as \mbox{A-0}
brings a green part to the green bin, so that the red bin comes to lie under
the gripper. The arm, already on its way down, moves over to the green bin
in one continuous motion and drops the part into it. Later in the same run
the person swaps the green and yellow bins, and \mbox{A-0} again puts the next
green part into the green bin at its new place. Two parts of the system
meet in this response: the policy places each part by what it sees, not by
where a bin was, and non-uniform streaming decoding keeps the motion
continuous as each new observation updates it (Section~\ref{sec:stream}).

\subsection{Other robots}
\label{sec:robots}

Nothing in \mbox{A-0} refers to the arm that carries the grippers: state and
action are defined at the grippers themselves
(Section~\ref{sec:anchored}), and the model sees only through the wrist
cameras. A robot that takes the YUBI gripper and its cameras as its end
effector can therefore run \mbox{A-0}. Besides the bimanual Franka, three
robots do: TX-G2, a humanoid; OpenArm, an open-source bimanual robot; and a
pair of UR industrial arms. All four run the same weights for each task, with
nothing retrained or adjusted for any robot.

\begin{figure}[!htb]
  \centering
  \input{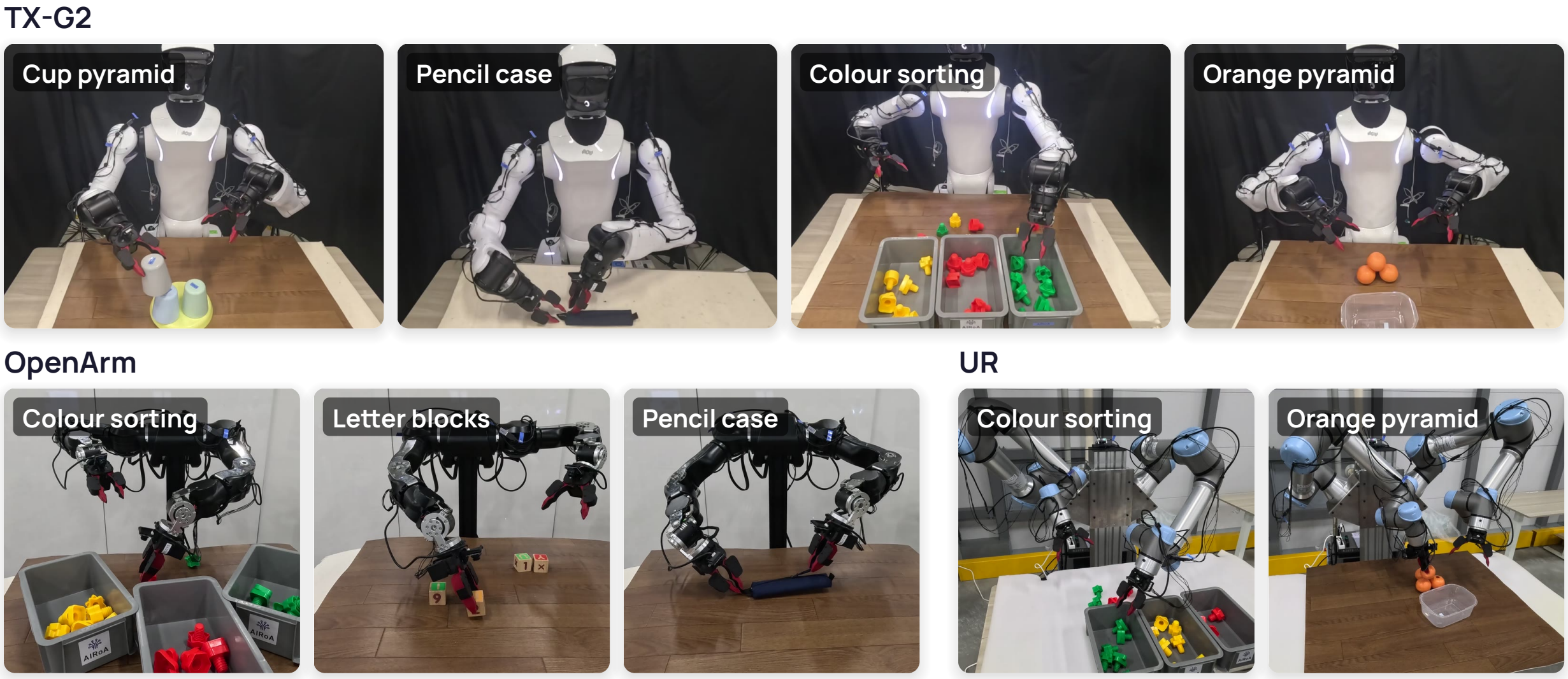}
  \caption{\mbox{A-0} on TX-G2, OpenArm and UR, each task in a frame from a run on
  that robot.}
  \label{fig:robots}
\end{figure}

On TX-G2, \mbox{A-0} performs four of the tasks of Section~\ref{sec:bench}: it stacks three cups
and four oranges into pyramids, opens the pencil case and sorts the toy
bolts and nuts into boxes by colour (Figure~\ref{fig:robots}, top). On
OpenArm, it performs three: it sorts the toy bolts and nuts by colour,
spells YUBI with the letter blocks and opens the pencil case; on the UR arms,
it sorts by colour and stacks four oranges into a pyramid
(Figure~\ref{fig:robots}, bottom). As on the Franka, none of the training data
was recorded on any of these robots.

\FloatBarrier

\section{Data}
\label{sec:data}

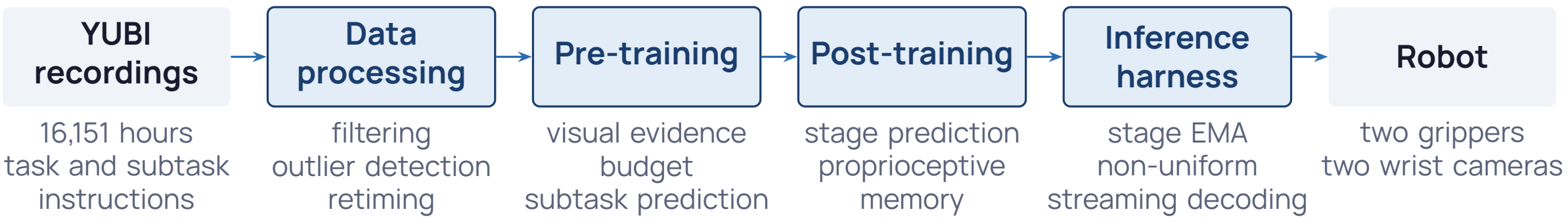
\begin{figure}[H]
  \centering
  \begin{tikzpicture}[
      box/.style={draw=Primary, line width=0.6pt, rounded corners=2pt,
        fill=AccentLight, minimum height=1.0cm, text width=2.15cm,
        align=center, inner sep=2pt, text=Primary,
        font=\sffamily\bfseries\footnotesize},
      edge/.style={rounded corners=2pt, fill=WarmBG, minimum height=1.0cm,
        text width=2.15cm, align=center, inner sep=2pt, text=TextBody,
        font=\sffamily\bfseries\footnotesize},
      note/.style={text width=2.5cm, align=center, text=TextMuted,
        inner sep=1pt, font=\sffamily\scriptsize},
      arr/.style={-{Stealth[length=4pt]}, draw=Accent, line width=0.6pt},
      node distance=0.36cm]
    \node[edge] (data) {YUBI\\recordings};
    \node[box, right=of data] (proc) {Data\\processing};
    \node[box, right=of proc] (pre) {Pre-training};
    \node[box, right=of pre] (post) {Post-training};
    \node[box, right=of post] (inf) {Inference\\harness};
    \node[edge, right=of inf] (robot) {Robot};
    \foreach \a/\b in {data/proc, proc/pre, pre/post, post/inf, inf/robot}
      \draw[arr] (\a) -- (\b);
    \node[note, below=3pt of data] {16,151 hours\\task and subtask\\instructions};
    \node[note, below=3pt of proc] {filtering\\outlier detection\\retiming};
    \node[note, below=3pt of pre] {visual evidence\\budget\\subtask prediction};
    \node[note, below=3pt of post] {stage prediction\\proprioceptive\\memory};
    \node[note, text width=2.9cm, below=3pt of inf] {stage EMA\\non-uniform\\streaming decoding};
    \node[note, below=3pt of robot] {two grippers\\two wrist cameras};
  \end{tikzpicture}
  \caption{\mbox{A-0} as one system. Handheld demonstrations pass through data
  processing into pre-training; post-training adapts the pre-trained model to
  the tasks it will perform; the inference harness runs it on the robot.}
  \label{fig:pipeline}
\end{figure}

The rest of this report describes how \mbox{A-0} is built, in four parts
designed with one another in view (Figure~\ref{fig:pipeline}). This section
describes the data, and the sections that follow take the other parts in
turn.

\subsection{YUBI}

YUBI~\cite{yubi} is a handheld gripper built for collecting manipulation
data at scale. It is driven by the fingers, so that closing the hand closes
the jaws; it carries a camera at the wrist; and it is tracked in six degrees
of freedom. A demonstrator holds one gripper in each hand and performs the
task as they naturally would, and the recording holds the two wrist views,
the trajectory of each gripper and how far each is open. The same gripper
mounts on a robot arm as its end effector. What a demonstrator records is
therefore what the robot will later see and do: the images come from the same
camera in the same place on the tool, and the actions are motions of that
tool.

\subsection{The corpus}

The corpus holds 16,151 hours of bimanual demonstrations
(Table~\ref{tab:corpus}). It covers well over a hundred tasks, spanning
pick-and-place, sorting, assembly, insertion, cable routing, opening and
closing, tool use and handovers between hands.

Every demonstration is divided into subtasks, steps of a few seconds each,
and it carries language at two levels: a \keyword{task instruction}, which
names the whole job, and a \keyword{subtask instruction}, which names the
step in progress. For one task, replacing the parts of a robot
vacuum, the task instruction is ``Replace the parts of the robot vacuum
cleaner'', and its subtasks run from ``Remove the first used mop pad and
place it into the box'' through ``Remove the action brush and place it into
the box'' to ``Attach the dustbin to the robot vacuum''. The corpus contains
10.65 million subtask segments.

These two levels of language run through the whole system. The task
instruction is what \mbox{A-0} is given at run time. The subtask instruction is
what it learns to predict in pre-training, and the unit from which the stages
of post-training are built. Before any training, every demonstration passes
through a data-processing pipeline that includes filtering, outlier detection
and retiming.

\begin{table}[H]
  \centering
  \caption{The corpus.}
  \label{tab:corpus}
  \vspace{0.5em}
  \small
  \begin{tabular}{@{}l l@{}}
    \toprule
    Demonstrations      & 16,151 hours, bimanual, recorded with YUBI \\
    Subtask segments    & 10.65 million, 5.5 seconds on average \\
    Time steps          & 1.74 billion at 30\,Hz \\
    Views               & two wrist cameras \\
    Proprioception      & pose and opening of each gripper \\
    Language            & a task instruction and a subtask instruction for every segment \\
    \bottomrule
  \end{tabular}
\end{table}

\section{Pre-training}
\label{sec:pretraining}

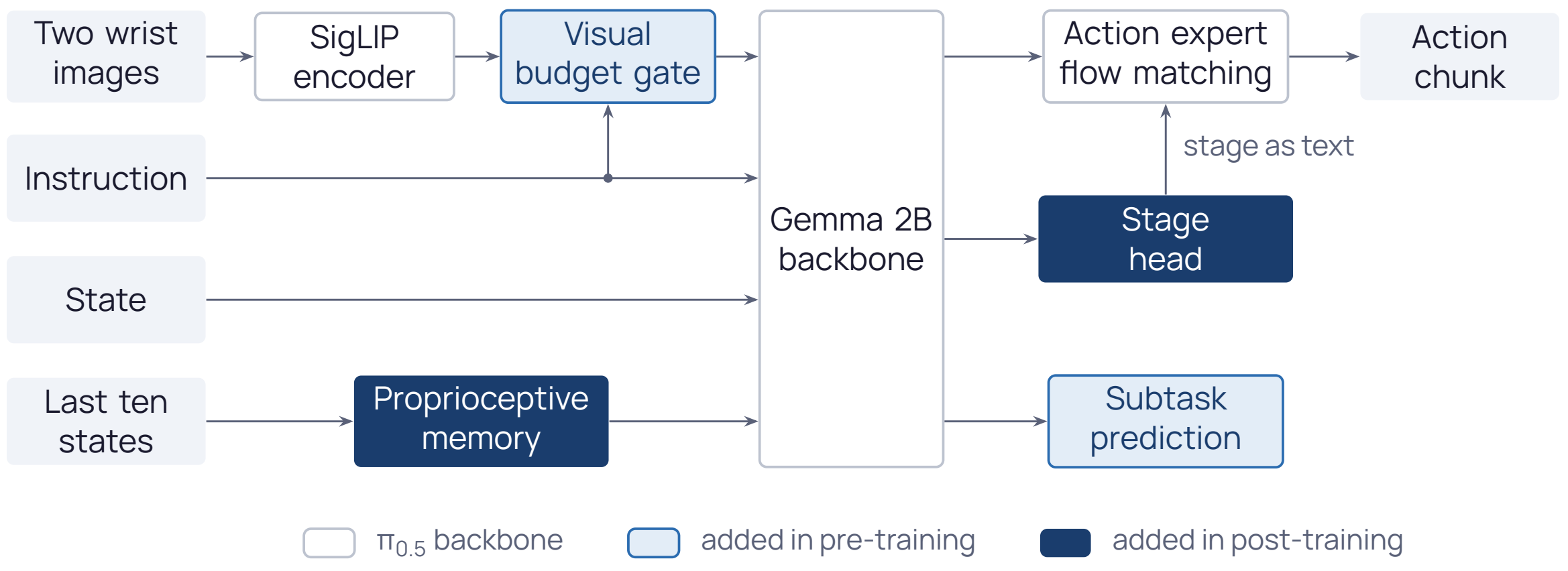
\begin{figure}[H]
  \centering
  \begin{tikzpicture}[
      base/.style={rounded corners=2pt, minimum height=0.86cm, align=center,
        inner sep=3pt, font=\sffamily\footnotesize},
      io/.style={base, fill=WarmBG, text=TextBody, text width=1.75cm},
      bb/.style={base, fill=white, draw=RuleGray, line width=0.7pt,
        text=TextBody, text width=1.75cm},
      pre/.style={base, fill=AccentLight, draw=Accent, line width=0.7pt,
        text=Primary, text width=1.9cm},
      post/.style={base, fill=Primary, text=white, text width=2.3cm},
      arr/.style={-{Stealth[length=4pt]}, draw=TextMuted, line width=0.55pt},
      lab/.style={font=\sffamily\scriptsize, text=TextMuted}]
    \node[io] (img)  at (0, 1.05)  {Two wrist\\images};
    \node[io] (ins)  at (0,-0.15)  {Instruction};
    \node[io] (st)   at (0,-1.35)  {State};
    \node[io] (hist) at (0,-2.55)  {Last ten\\states};
    \node[bb]   (enc)  at (2.45, 1.05) {SigLIP\\encoder};
    \node[pre]  (gate) at (4.95, 1.05) {Visual\\budget gate};
    \node[post] (mem)  at (3.7,-2.55)  {Proprioceptive\\memory};
    \node[bb, minimum height=4.5cm, text width=1.6cm] (llm) at (7.35,-0.75)
      {Gemma 2B\\backbone};
    \node[bb, text width=2.2cm] (exp) at (10.45, 1.05) {Action expert\\flow matching};
    \node[io] (act) at (13.35, 1.05) {Action\\chunk};
    \node[post] (stage) at (10.45,-0.75) {Stage\\head};
    \node[pre, text width=2.1cm] (sub) at (10.45,-2.55) {Subtask\\prediction};
    \draw[arr] (img) -- (enc);
    \draw[arr] (enc) -- (gate);
    \draw[arr] (gate) -- (gate -| llm.west);
    \draw[arr] (ins) -- (ins -| llm.west);
    \draw[arr] (st) -- (st -| llm.west);
    \draw[arr] (hist) -- (mem);
    \draw[arr] (mem) -- (mem -| llm.west);
    \fill[TextMuted] (4.95,-0.15) circle (1.3pt);
    \draw[arr] (4.95,-0.15) -- (gate.south);
    \draw[arr] (llm.east |- exp) -- (exp);
    \draw[arr] (exp) -- (act);
    \draw[arr] (llm.east |- stage) -- (stage);
    \draw[arr] (stage) -- node[lab, right=1pt] {stage as text} (exp);
    \draw[arr] (llm.east |- sub) -- (sub);
    \node[bb, minimum height=0.28cm, minimum width=0.5cm, text width=0.3cm, inner sep=0pt] (l1) at (2.2,-3.75) {};
    \node[lab, right=2pt of l1] (l1t) {π\textsubscript{0.5} backbone};
    \node[pre, minimum height=0.28cm, minimum width=0.5cm, text width=0.3cm, inner sep=0pt, right=14pt of l1t] (l2) {};
    \node[lab, right=2pt of l2] (l2t) {added in pre-training};
    \node[post, minimum height=0.28cm, minimum width=0.5cm, text width=0.3cm, inner sep=0pt, right=14pt of l2t] (l3) {};
    \node[lab, right=2pt of l3] {added in post-training};
  \end{tikzpicture}
  \caption{The \mbox{A-0} model. Pre-training adds the visual budget gate and
  subtask prediction to the \pio{} backbone. Post-training adds the stage
  head, whose prediction conditions the action expert as text, and the
  proprioceptive memory.}
  \label{fig:model}
\end{figure}

Pre-training teaches one model the corpus under the objective of See Less,
Specify More (S2)~\cite{s2}, which governs what the policy sees and how it is
told what to do, with an auxiliary loss on the subtask
(Figure~\ref{fig:model}).

\subsection{Model}

\mbox{A-0} takes its backbone from \pio{}~\cite{pi05}: a SigLIP So400m/14 image
encoder, a Gemma~2B vision-language backbone and a 300M action expert
trained by flow matching, initialised from the public \pio{} base
checkpoint. On this backbone \mbox{A-0} redefines the interface to the robot,
adds a visual budget gate and changes the objective, and backbone and gate are
trained together on the corpus for 300,000 steps at a batch size of 2,048.

The model observes two wrist images at $224\times224$ pixels, an instruction
and a proprioceptive state, and in pre-training it outputs a chunk of 32 future
actions on a 20\,Hz grid.

\Needspace{9\baselineskip}
\subsection{Anchored actions}
\label{sec:anchored}

State and action are both defined at the grippers. Actions are
anchored: every action in a chunk is the pose a gripper should
reach, expressed relative to the pose it has when the observation is taken.
With $T_i \in SE(3)$ the pose of a gripper's tool centre point at time step
$i$, the $k$-th of the $H$ actions of the chunk predicted at $i$ is
\begin{equation}
  a_k = T_i^{-1}\, T_{i+k}, \qquad k = 1, \dots, H,
  \label{eq:anchored}
\end{equation}
for each arm, written as a translation and a 6D rotation, together with the
gripper command at $i+k$: 20 numbers per action. The state is the pose of
one gripper in the frame of the other, with the two gripper openings: 11
numbers.

The representation follows the relative trajectories of UMI~\cite{umi}, and
it lets handheld and robot trajectories share the same coordinates. No quantity refers to a
robot base, a joint or a world frame, none of which a person holding the
grippers and a robot carrying them share. The same numbers describe that person and a robot
carrying the grippers, so handheld data trains a robot policy directly and
one policy can drive different arms. And because every action is a target in its
own right, not an increment on the one before it, errors do not accumulate
along the chunk, and the robot's controller can follow the chunk at its own
rate.

\subsection{Language at two levels}

S2 pairs a restriction on what the policy sees (See Less) with language that
specifies the step being executed as well as the goal (Specify More), and
\mbox{A-0} uses both. On the language side, each training sample is conditioned on
either the task instruction or the subtask instruction, so that the
pre-trained model can be driven at either level, by one instruction for the
whole task or step by step, and this is what later allows a predicted stage
to steer it.

\subsection{Visual evidence budget}

\begin{figure}[t]
  \centering
  \setlength{\tabcolsep}{1.6pt}\renewcommand{\arraystretch}{1}%
  {\sffamily\fontsize{6.5}{8}\selectfont\color{TextMuted}%
  \begin{tabular}{@{}ccccc@{}}
    \gatetile{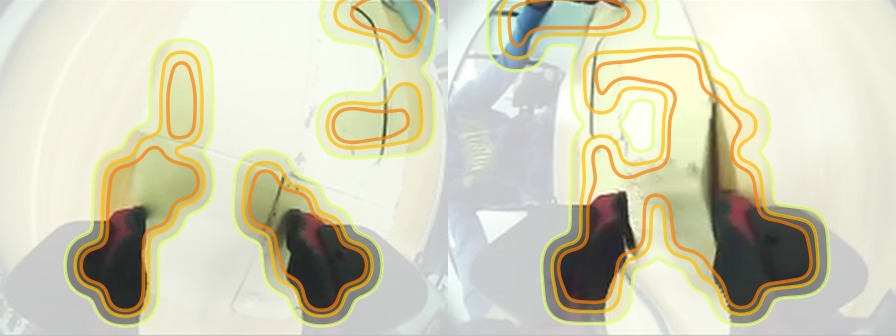}{Tissue box: turn over} &
    \gatetile{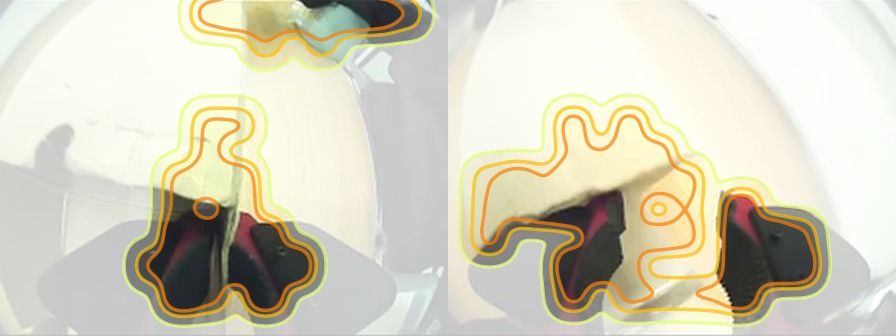}{Tissue box: fold flap} &
    \gatetile{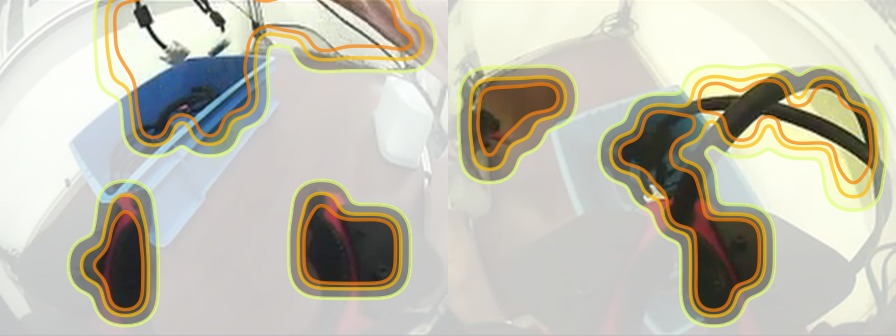}{Pick up DVI cable} &
    \gatetile{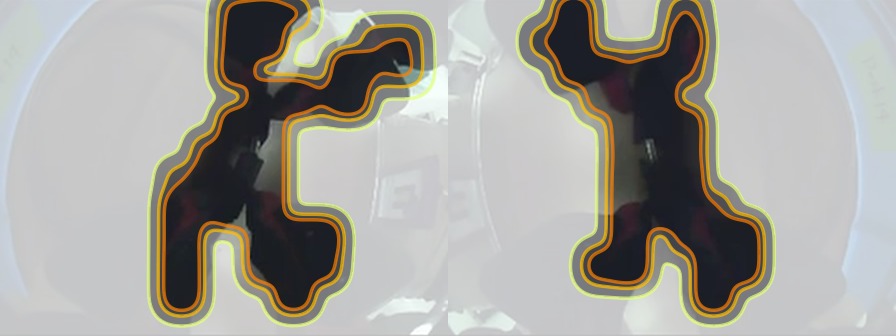}{Connect cable} &
    \gatetile{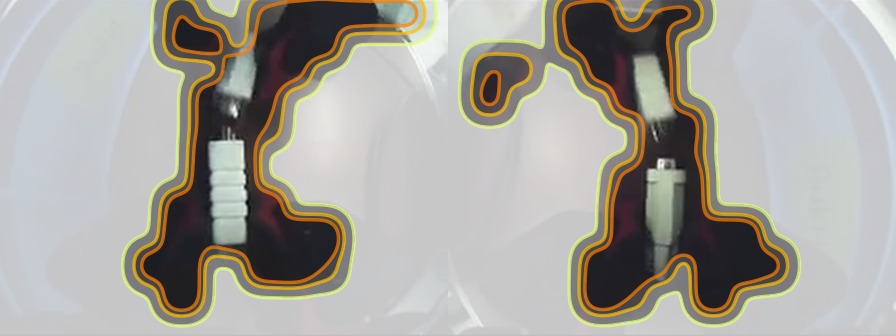}{Disconnect cable} \\\noalign{\vskip 5pt}
    \gatetile{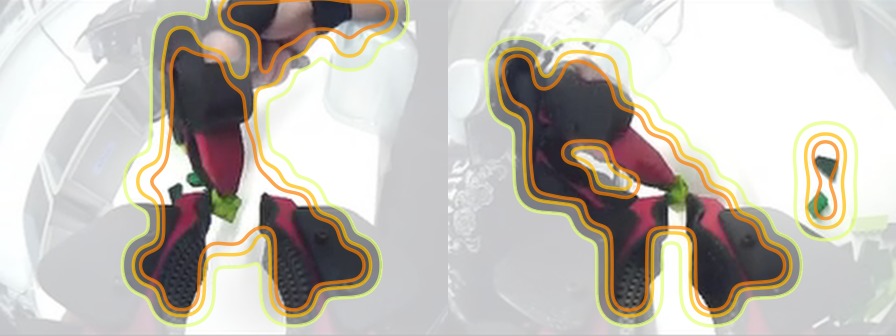}{Pick LEGO brick} &
    \gatetile{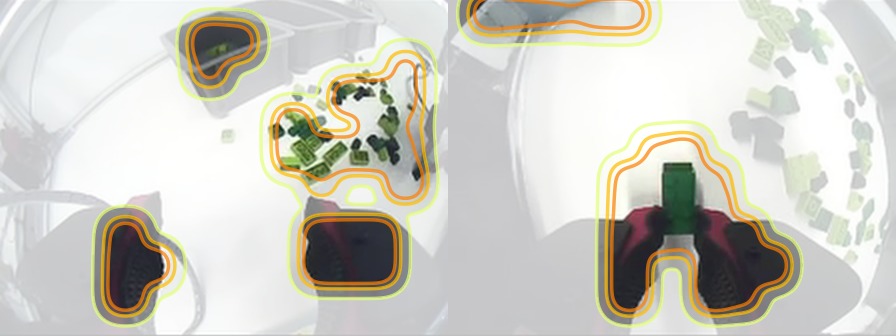}{Stack LEGO bricks} &
    \gatetile{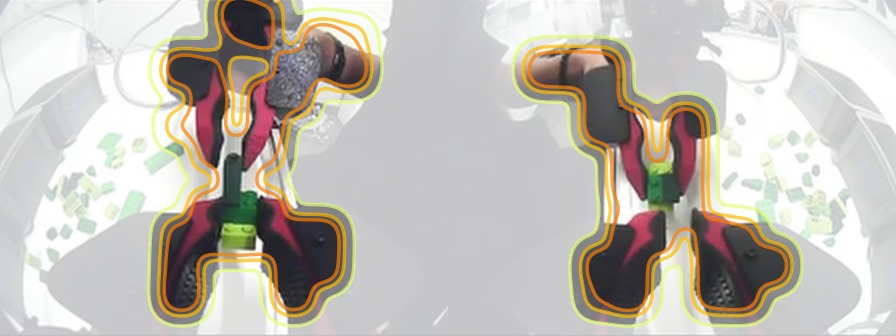}{Toss LEGO by colour} &
    \gatetile{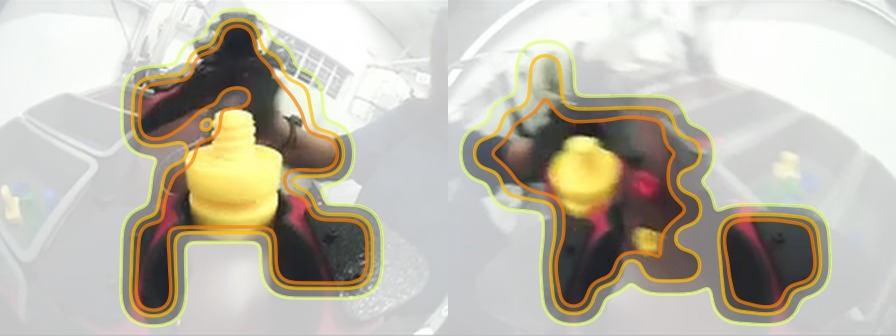}{Separate bolt and nut} &
    \gatetile{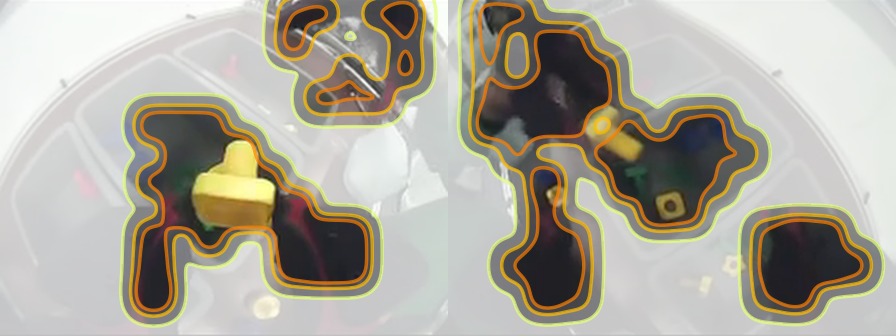}{Sort bolt and nut} \\\noalign{\vskip 5pt}
    \gatetile{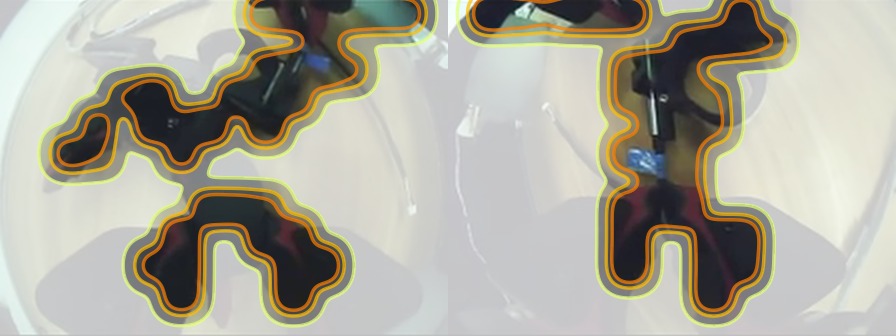}{Insert USB-C into YUBI} &
    \gatetile{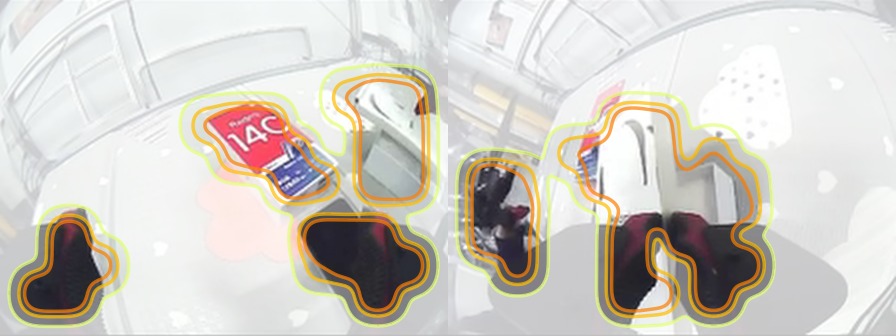}{Insert phone-box divider} &
    \gatetile{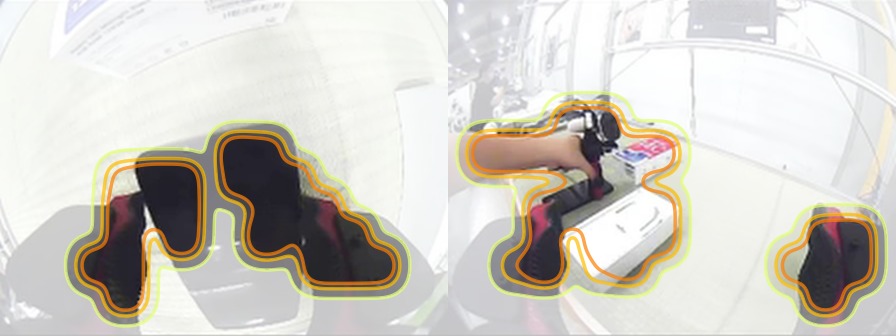}{Stand smartphone upright} &
    \gatetile{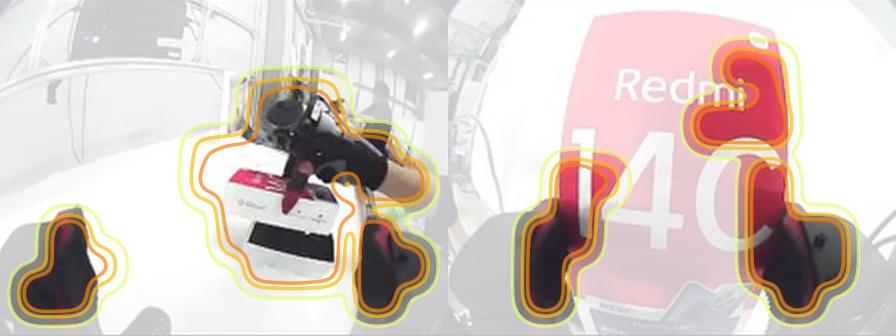}{Close smartphone box} &
    \gatetile{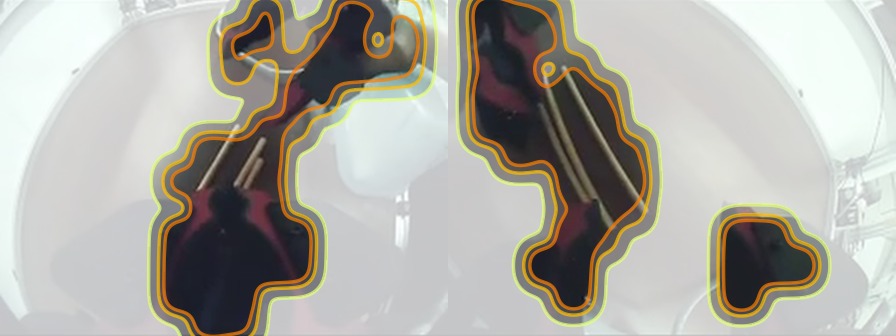}{Unzip pencil case} \\\noalign{\vskip 5pt}
    \gatetile{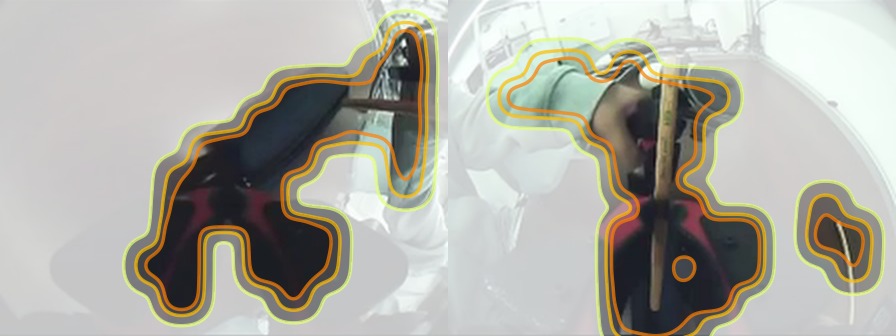}{Put pencil into case} &
    \gatetile{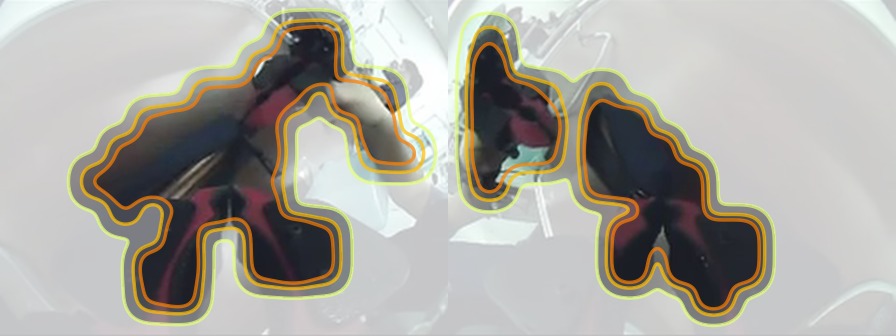}{Zip full pencil case} &
    \gatetile{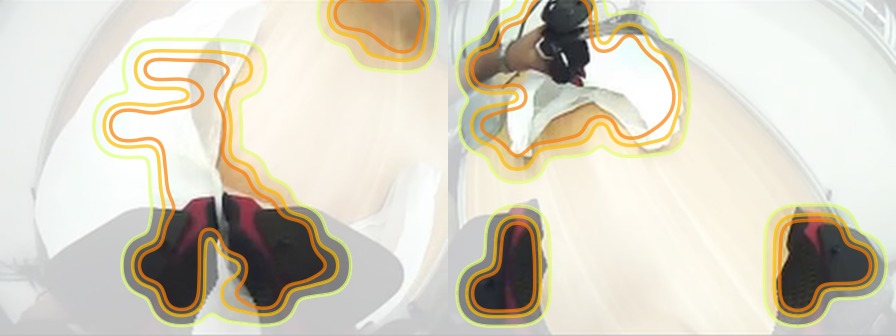}{Open plastic bag} &
    \gatetile{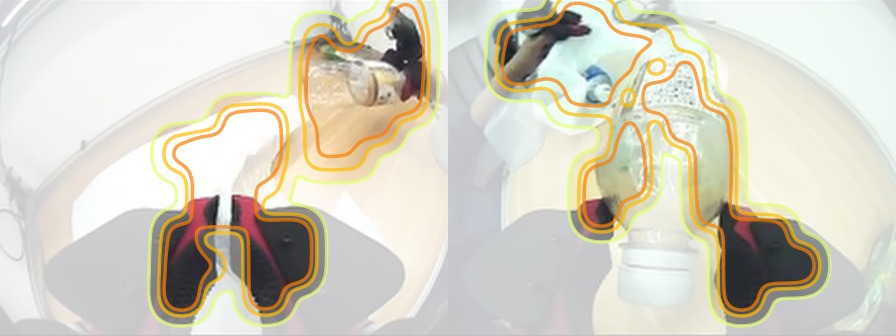}{Bottle into plastic bag} &
    \gatetile{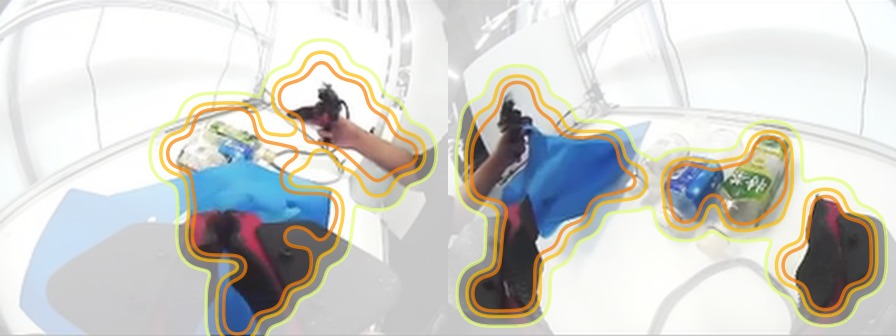}{Open tote bag} \\\noalign{\vskip 5pt}
    \gatetile{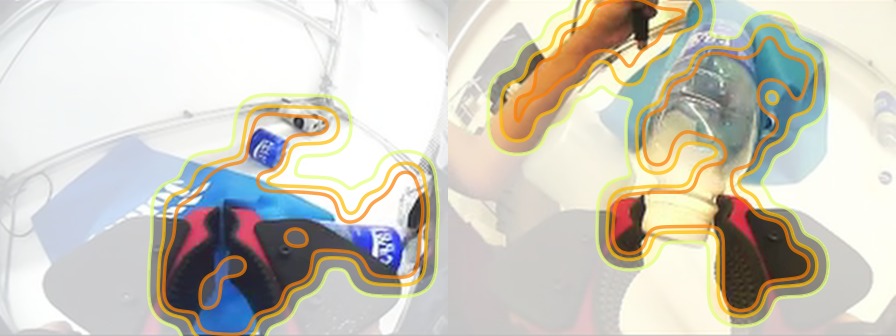}{Bottle out of tote bag} &
    \gatetile{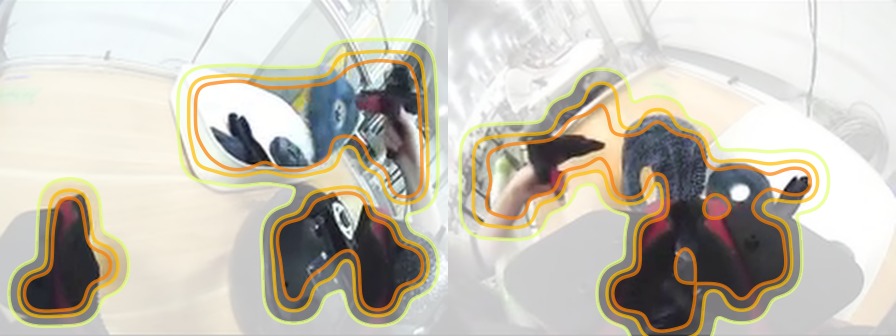}{Attach robot-vacuum mop} &
    \gatetile{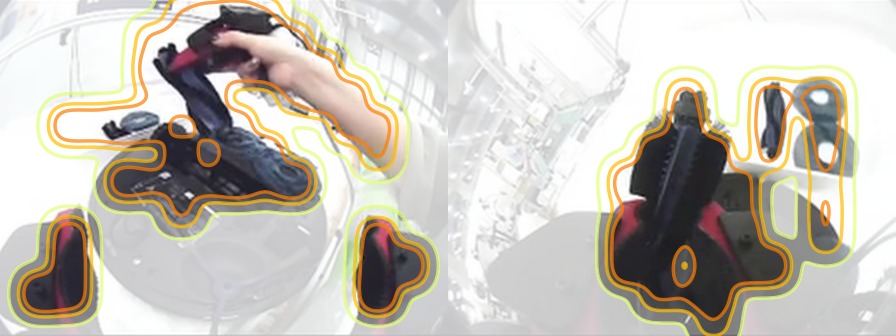}{Remove vacuum brush} &
    \gatetile{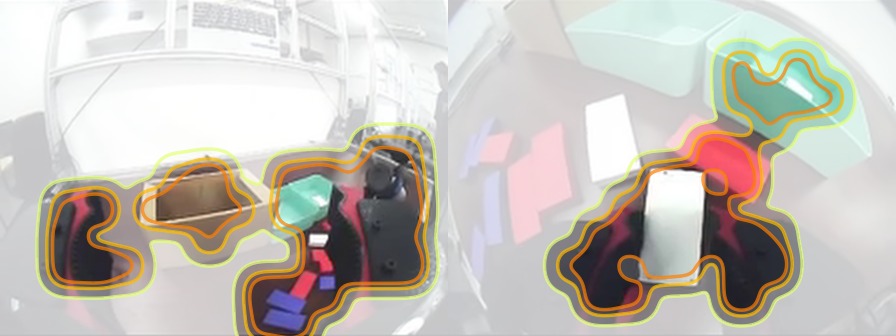}{Sort domino by colour} &
    \gatetile{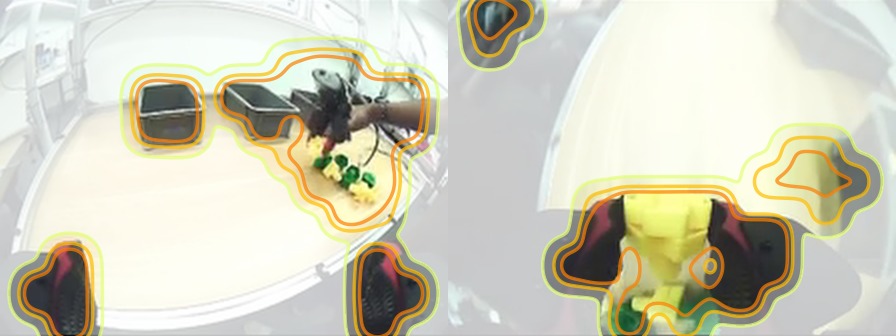}{Sort toy bolt by colour}
  \end{tabular}}
  \caption{Where the visual budget gate puts its weight, on 25 subtasks. Each tile shows the left and right wrist views the model
  receives. The contours trace the smoothed keep probability at three levels,
  and the view fades where the gate gives it less weight.}
  \label{fig:gate}
\end{figure}

A wrist camera shows far more than a task needs, and what it shows beyond the
task is what changes from one site to the next. S2 lets the policy learn
what to look at. The visual budget gate scores every image token against the
instruction: for token $z_p$ of a view and a pooled summary $q$ of the
prompt, a small network $h$ returns a keep probability $m_p$, and the token
that reaches the backbone is scaled by it,
\begin{equation}
  m_p = \sigma\bigl(h(z_p,\, q,\, z_p \odot q)\,/\,\tau\bigr),
  \qquad
  \hat{z}_p = \bigl(\gamma + (1-\gamma)\, m_p\bigr)\, z_p ,
  \label{eq:gate}
\end{equation}
where $\tau$ is a temperature and $\gamma$ a floor that keeps a trace of
every token. A budget loss holds the mean keep probability over the $P$
tokens of a view to a target $\rho$:
\begin{equation}
  \Lc_{\text{budget}} = \Bigl(\frac{1}{P}\sum_{p=1}^{P} m_p - \rho\Bigr)^{2}.
  \label{eq:budget}
\end{equation}
The budget constrains the average gate weight, while the action loss teaches
the gate which regions to emphasise when the model decides how to act,
without masks, boxes or labels indicating where to look. Across 25 subtasks (Figure~\ref{fig:gate}) the gate favours the grippers and the
objects they are working on, while the table, the background and the rest of
the room recede.

\FloatBarrier
\subsection{Subtask auxiliary loss}

When the prompt is the task instruction, the model is also asked to name the
step in progress: from the same observation, the language backbone predicts
the subtask instruction as text, with a loss $\Lc_{\text{sub}}$. The
prediction is used only in training. It is never an input to the action
expert, and its purpose is to shape the representation: to name the current
step from two wrist views and a task-level instruction, the backbone has to
encode where the task stands, and the action expert reads that same
representation.

\subsection{Scaling with pre-training data}
\label{sec:scaling}

\begin{figure}[t]
  \centering
  \begin{minipage}[t]{0.5\linewidth}\raggedright
    \input{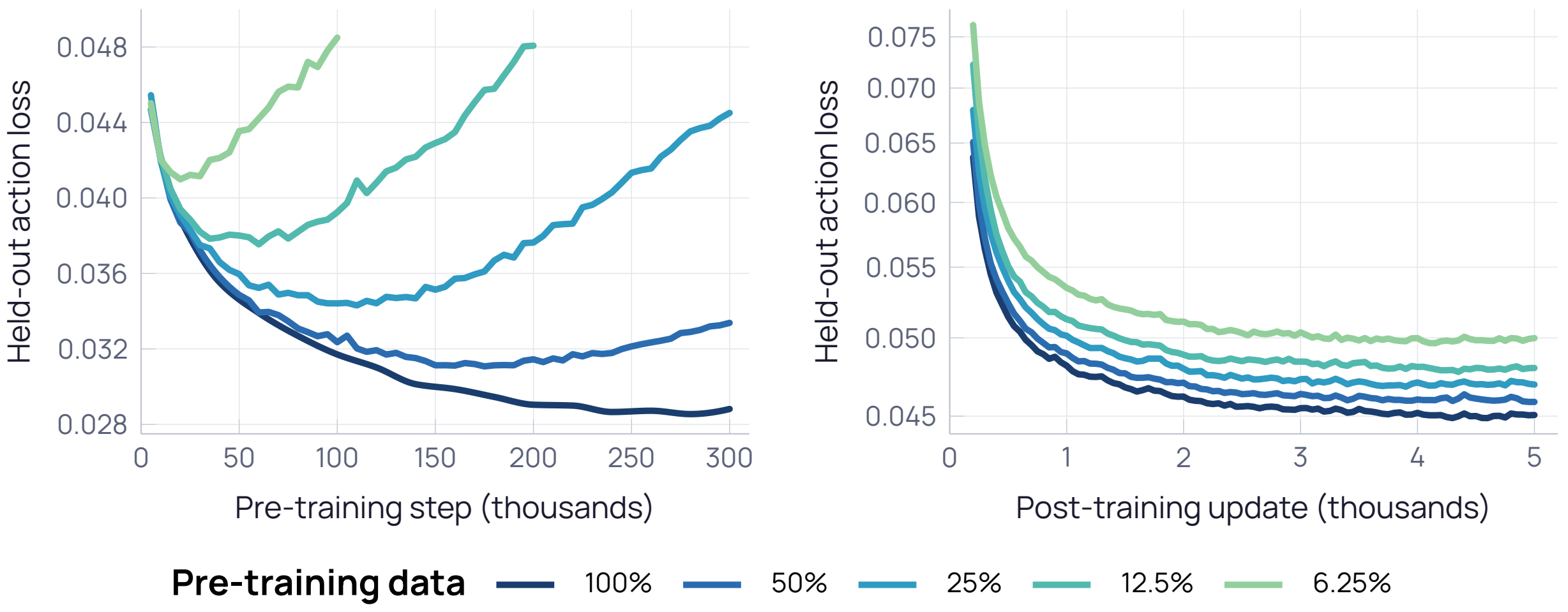}\unskip
  \end{minipage}%
  \begin{minipage}[t]{0.5\linewidth}\raggedleft
    \definecolor{ScaleA}{HTML}{1A3D6D}
\definecolor{ScaleB}{HTML}{2B6CB0}
\definecolor{ScaleC}{HTML}{2F9AC2}
\definecolor{ScaleD}{HTML}{52B9AE}
\definecolor{ScaleE}{HTML}{93CF9C}
\begin{tikzpicture}
  \begin{axis}[
      scale only axis, width=6.3cm, height=4.4cm,
      xlabel={Post-training update (thousands)},
      ylabel={Held-out action loss},
      ymin=0.04394, ymax=0.07784, ytick={0.045,0.050,0.055,0.060,0.065,0.070,0.075}, yticklabels={0.045,0.050,0.055,0.060,0.065,0.070,0.075},
      tick label style={font=\sffamily\scriptsize, text=TextMuted},
      label style={font=\sffamily\footnotesize, text=TextBody},
      axis line style={RuleGray}, tick style={RuleGray},
      ymajorgrids, xmajorgrids, grid style={RuleGray!40, line width=0.3pt},
      axis x line*=bottom, axis y line*=left,
      ymode=log, minor ytick={},
      xmin=0, xmax=5.2, xtick={0,1,2,3,4,5},
    ]
  \addplot[ScaleA, line width=1.9pt, line join=round, line cap=round, no markers] coordinates {(0.2,0.063769) (0.25,0.058886) (0.3,0.056362) (0.35,0.054498) (0.4,0.053138) (0.45,0.052271) (0.5,0.051439) (0.55,0.050803) (0.6,0.050424) (0.65,0.049868) (0.7,0.049519) (0.75,0.049102) (0.8,0.048913) (0.85,0.048611) (0.9,0.048744) (0.95,0.048391) (1,0.048163) (1.05,0.047814) (1.1,0.047610) (1.15,0.047579) (1.2,0.047456) (1.25,0.047440) (1.3,0.047476) (1.35,0.047222) (1.4,0.047050) (1.45,0.046971) (1.5,0.046773) (1.55,0.046669) (1.6,0.046516) (1.65,0.046603) (1.7,0.046726) (1.75,0.046571) (1.8,0.046546) (1.85,0.046376) (1.9,0.046234) (1.95,0.046195) (2,0.046201) (2.05,0.046004) (2.1,0.045943) (2.15,0.045921) (2.2,0.045790) (2.25,0.045741) (2.3,0.045783) (2.35,0.045640) (2.4,0.045747) (2.45,0.045572) (2.5,0.045589) (2.55,0.045614) (2.6,0.045568) (2.65,0.045478) (2.7,0.045562) (2.75,0.045580) (2.8,0.045551) (2.85,0.045416) (2.9,0.045407) (2.95,0.045385) (3,0.045486) (3.05,0.045494) (3.1,0.045396) (3.15,0.045459) (3.2,0.045279) (3.25,0.045319) (3.3,0.045457) (3.35,0.045369) (3.4,0.045207) (3.45,0.045259) (3.5,0.045441) (3.55,0.045314) (3.6,0.045283) (3.65,0.045225) (3.7,0.045147) (3.75,0.045115) (3.8,0.045195) (3.85,0.045229) (3.9,0.045089) (3.95,0.045096) (4,0.045200) (4.05,0.045093) (4.1,0.045055) (4.15,0.045053) (4.2,0.045067) (4.25,0.044946) (4.3,0.044884) (4.35,0.044991) (4.4,0.045164) (4.45,0.044993) (4.5,0.045005) (4.55,0.044890) (4.6,0.044883) (4.65,0.045010) (4.7,0.045021) (4.75,0.044918) (4.8,0.045159) (4.85,0.045103) (4.9,0.045099) (4.95,0.045055) (5,0.045068)};
  \addplot[ScaleB, line width=1.9pt, line join=round, line cap=round, no markers] coordinates {(0.2,0.065083) (0.25,0.059972) (0.3,0.057204) (0.35,0.055294) (0.4,0.054107) (0.45,0.053214) (0.5,0.052392) (0.55,0.051666) (0.6,0.051170) (0.65,0.050619) (0.7,0.050375) (0.75,0.049955) (0.8,0.049663) (0.85,0.049366) (0.9,0.049386) (0.95,0.049060) (1,0.048969) (1.05,0.048672) (1.1,0.048445) (1.15,0.048466) (1.2,0.048303) (1.25,0.048291) (1.3,0.048236) (1.35,0.048027) (1.4,0.047900) (1.45,0.047688) (1.5,0.047670) (1.55,0.047447) (1.6,0.047383) (1.65,0.047387) (1.7,0.047420) (1.75,0.047332) (1.8,0.047266) (1.85,0.047104) (1.9,0.047113) (1.95,0.047045) (2,0.047070) (2.05,0.046852) (2.1,0.046833) (2.15,0.046680) (2.2,0.046647) (2.25,0.046537) (2.3,0.046555) (2.35,0.046430) (2.4,0.046484) (2.45,0.046449) (2.5,0.046352) (2.55,0.046376) (2.6,0.046413) (2.65,0.046325) (2.7,0.046307) (2.75,0.046359) (2.8,0.046411) (2.85,0.046281) (2.9,0.046255) (2.95,0.046227) (3,0.046368) (3.05,0.046305) (3.1,0.046228) (3.15,0.046252) (3.2,0.046109) (3.25,0.046121) (3.3,0.046269) (3.35,0.046139) (3.4,0.045973) (3.45,0.046068) (3.5,0.046248) (3.55,0.046192) (3.6,0.046110) (3.65,0.046078) (3.7,0.045971) (3.75,0.046053) (3.8,0.045998) (3.85,0.045995) (3.9,0.045949) (3.95,0.045946) (4,0.046019) (4.05,0.045984) (4.1,0.045987) (4.15,0.045979) (4.2,0.046096) (4.25,0.046142) (4.3,0.045927) (4.35,0.046077) (4.4,0.046343) (4.45,0.046139) (4.5,0.046078) (4.55,0.046023) (4.6,0.045980) (4.65,0.045948) (4.7,0.045967) (4.75,0.046011) (4.8,0.046186) (4.85,0.046095) (4.9,0.045908) (4.95,0.045870) (5,0.045870)};
  \addplot[ScaleC, line width=1.9pt, line join=round, line cap=round, no markers] coordinates {(0.2,0.067959) (0.25,0.062540) (0.3,0.059601) (0.35,0.057597) (0.4,0.056095) (0.45,0.054974) (0.5,0.054017) (0.55,0.053185) (0.6,0.052680) (0.65,0.052096) (0.7,0.051720) (0.75,0.051287) (0.8,0.050934) (0.85,0.050620) (0.9,0.050519) (0.95,0.050242) (1,0.050144) (1.05,0.049869) (1.1,0.049696) (1.15,0.049506) (1.2,0.049320) (1.25,0.049324) (1.3,0.049330) (1.35,0.049115) (1.4,0.048978) (1.45,0.048753) (1.5,0.048648) (1.55,0.048569) (1.6,0.048434) (1.65,0.048475) (1.7,0.048629) (1.75,0.048621) (1.8,0.048627) (1.85,0.048309) (1.9,0.048159) (1.95,0.048168) (2,0.047968) (2.05,0.047774) (2.1,0.047785) (2.15,0.047815) (2.2,0.047550) (2.25,0.047551) (2.3,0.047494) (2.35,0.047383) (2.4,0.047571) (2.45,0.047449) (2.5,0.047333) (2.55,0.047408) (2.6,0.047280) (2.65,0.047291) (2.7,0.047338) (2.75,0.047390) (2.8,0.047349) (2.85,0.047176) (2.9,0.047232) (2.95,0.047162) (3,0.047294) (3.05,0.047311) (3.1,0.047025) (3.15,0.047156) (3.2,0.047089) (3.25,0.047166) (3.3,0.047321) (3.35,0.047167) (3.4,0.046975) (3.45,0.047084) (3.5,0.047155) (3.55,0.047100) (3.6,0.046997) (3.65,0.046910) (3.7,0.046866) (3.75,0.046899) (3.8,0.046904) (3.85,0.046925) (3.9,0.046820) (3.95,0.046999) (4,0.047091) (4.05,0.046981) (4.1,0.046885) (4.15,0.046896) (4.2,0.047044) (4.25,0.046964) (4.3,0.046943) (4.35,0.047074) (4.4,0.047146) (4.45,0.047022) (4.5,0.046967) (4.55,0.046982) (4.6,0.046938) (4.65,0.046956) (4.7,0.047032) (4.75,0.046902) (4.8,0.047186) (4.85,0.047109) (4.9,0.047094) (4.95,0.046999) (5,0.046955)};
  \addplot[ScaleD, line width=1.9pt, line join=round, line cap=round, no markers] coordinates {(0.2,0.072241) (0.25,0.065699) (0.3,0.061979) (0.35,0.059424) (0.4,0.057471) (0.45,0.056186) (0.5,0.055079) (0.55,0.054308) (0.6,0.053941) (0.65,0.053169) (0.7,0.052870) (0.75,0.052421) (0.8,0.052139) (0.85,0.051772) (0.9,0.051780) (0.95,0.051480) (1,0.051269) (1.05,0.051151) (1.1,0.050841) (1.15,0.050745) (1.2,0.050662) (1.25,0.050607) (1.3,0.050570) (1.35,0.050299) (1.4,0.050153) (1.45,0.049975) (1.5,0.049841) (1.55,0.049730) (1.6,0.049710) (1.65,0.049547) (1.7,0.049578) (1.75,0.049528) (1.8,0.049486) (1.85,0.049187) (1.9,0.049062) (1.95,0.049002) (2,0.048880) (2.05,0.048757) (2.1,0.048755) (2.15,0.048787) (2.2,0.048586) (2.25,0.048501) (2.3,0.048466) (2.35,0.048509) (2.4,0.048631) (2.45,0.048538) (2.5,0.048455) (2.55,0.048569) (2.6,0.048539) (2.65,0.048492) (2.7,0.048405) (2.75,0.048505) (2.8,0.048565) (2.85,0.048440) (2.9,0.048610) (2.95,0.048424) (3,0.048429) (3.05,0.048420) (3.1,0.048265) (3.15,0.048301) (3.2,0.048133) (3.25,0.048311) (3.3,0.048444) (3.35,0.048220) (3.4,0.048157) (3.45,0.048328) (3.5,0.048367) (3.55,0.048404) (3.6,0.048144) (3.65,0.048179) (3.7,0.048026) (3.75,0.047974) (3.8,0.047989) (3.85,0.048110) (3.9,0.047954) (3.95,0.048140) (4,0.048083) (4.05,0.048092) (4.1,0.048013) (4.15,0.047939) (4.2,0.047867) (4.25,0.047867) (4.3,0.047883) (4.35,0.047763) (4.4,0.047966) (4.45,0.047928) (4.5,0.048019) (4.55,0.048005) (4.6,0.047908) (4.65,0.047918) (4.7,0.047980) (4.75,0.047901) (4.8,0.048049) (4.85,0.048119) (4.9,0.047946) (4.95,0.048005) (5,0.048014)};
  \addplot[ScaleE, line width=1.9pt, line join=round, line cap=round, no markers] coordinates {(0.2,0.076207) (0.25,0.068723) (0.3,0.064872) (0.35,0.062261) (0.4,0.060465) (0.45,0.059220) (0.5,0.057974) (0.55,0.057160) (0.6,0.056550) (0.65,0.055786) (0.7,0.055480) (0.75,0.054995) (0.8,0.054645) (0.85,0.054282) (0.9,0.054110) (0.95,0.053798) (1,0.053480) (1.05,0.053294) (1.1,0.052994) (1.15,0.052929) (1.2,0.052700) (1.25,0.052588) (1.3,0.052677) (1.35,0.052309) (1.4,0.052135) (1.45,0.052025) (1.5,0.051954) (1.55,0.051880) (1.6,0.051699) (1.65,0.051625) (1.7,0.051638) (1.75,0.051570) (1.8,0.051630) (1.85,0.051217) (1.9,0.051147) (1.95,0.051090) (2,0.051112) (2.05,0.050928) (2.1,0.050936) (2.15,0.050884) (2.2,0.050603) (2.25,0.050609) (2.3,0.050389) (2.35,0.050442) (2.4,0.050535) (2.45,0.050445) (2.5,0.050321) (2.55,0.050172) (2.6,0.050422) (2.65,0.050294) (2.7,0.050249) (2.75,0.050306) (2.8,0.050430) (2.85,0.050289) (2.9,0.050328) (2.95,0.050145) (3,0.050373) (3.05,0.050200) (3.1,0.050025) (3.15,0.050113) (3.2,0.049958) (3.25,0.049935) (3.3,0.050224) (3.35,0.049861) (3.4,0.050015) (3.45,0.049944) (3.5,0.049825) (3.55,0.049990) (3.6,0.049785) (3.65,0.049959) (3.7,0.049871) (3.75,0.049953) (3.8,0.049848) (3.85,0.049798) (3.9,0.049776) (3.95,0.049967) (4,0.049984) (4.05,0.049770) (4.1,0.049655) (4.15,0.049641) (4.2,0.049773) (4.25,0.049821) (4.3,0.049759) (4.35,0.049829) (4.4,0.050076) (4.45,0.049920) (4.5,0.049843) (4.55,0.049843) (4.6,0.049810) (4.65,0.049797) (4.7,0.049975) (4.75,0.049805) (4.8,0.049985) (4.85,0.049958) (4.9,0.049805) (4.95,0.049928) (5,0.049985)};
  \end{axis}
\end{tikzpicture}\unskip
  \end{minipage}\par
  \vspace{4pt}
  \definecolor{ScaleA}{HTML}{1A3D6D}
\definecolor{ScaleB}{HTML}{2B6CB0}
\definecolor{ScaleC}{HTML}{2F9AC2}
\definecolor{ScaleD}{HTML}{52B9AE}
\definecolor{ScaleE}{HTML}{93CF9C}
\begin{tikzpicture}
  \begin{axis}[hide axis, scale only axis, width=1pt, height=1pt, xmin=0, xmax=1, ymin=0, ymax=1,
      legend columns=6, legend style={font=\sffamily\scriptsize, draw=none, fill=none,
        column sep=6pt, at={(0.5,1)}, anchor=north}]
    \addlegendimage{empty legend}\addlegendentry{\textbf{Pre-training data}}
    \addlegendimage{ScaleA, line width=1.9pt}\addlegendentry{100\%} \addlegendimage{ScaleB, line width=1.9pt}\addlegendentry{50\%} \addlegendimage{ScaleC, line width=1.9pt}\addlegendentry{25\%} \addlegendimage{ScaleD, line width=1.9pt}\addlegendentry{12.5\%} \addlegendimage{ScaleE, line width=1.9pt}\addlegendentry{6.25\%}
  \end{axis}
\end{tikzpicture}
  \caption{Action loss on held-out demonstrations for models pre-trained on
  a share of the data. Left, during pre-training, with the task instruction
  as prompt (the subtask instruction gives the same picture): every run has
  the same model, batch size and schedule, and the 100\% run is the
  pre-training of \mbox{A-0}. Right, during post-training, averaged over 16
  task sets: each run starts from its checkpoint at 100,000 pre-training
  steps, and post-training data and settings are the same for all.}
  \label{fig:scaling}
\end{figure}

Handheld data is collected so that it can be scaled, and what matters is
whether more of it makes a better model. We pre-train the same model on
50\%, 25\%, 12.5\% and 6.25\% of the pre-training data, at the batch size
and with the schedule of the full run, and compare their action loss on
3,072 action chunks from held-out recordings (Figure~\ref{fig:scaling}, left). Once training is under way, the
more data the model is pre-trained on, the lower its loss and the longer it
keeps improving, and the full corpus gives the lowest loss of all: recording
more handheld data pays off.

\begin{figure}[t]
  \centering
  \definecolor{ScaleA}{HTML}{1A3D6D}
\definecolor{ScaleB}{HTML}{2B6CB0}
\definecolor{ScaleC}{HTML}{2F9AC2}
\definecolor{ScaleD}{HTML}{52B9AE}
\definecolor{ScaleE}{HTML}{93CF9C}
\begin{tikzpicture}
  \begin{axis}[
      scale only axis, width=6.3cm, height=4.4cm,
      xlabel={Pre-training data (hours)},
      ylabel={Held-out action loss},
      xmode=log, ymode=log, minor xtick={}, minor ytick={},
      xmin=620, xmax=21000, ymin=0.04412, ymax=0.05087, ytick={0.045,0.046,0.047,0.048,0.049,0.050}, yticklabels={0.045,0.046,0.047,0.048,0.049,0.050},
      xtick={1009.44,2018.88,4037.75,8075.50,16151.00},
      xticklabels={1k,2k,4k,8k,16k},
      tick label style={font=\sffamily\scriptsize, text=TextMuted},
      label style={font=\sffamily\footnotesize, text=TextBody},
      axis line style={RuleGray}, tick style={RuleGray},
      ymajorgrids, xmajorgrids, grid style={RuleGray!40, line width=0.3pt},
      axis x line*=bottom, axis y line*=left,
      legend style={font=\sffamily\scriptsize, draw=none, fill=white, fill opacity=0.85,
        text opacity=1, at={(0.98,0.97)}, anchor=north east},
      clip=false,
    ]
  \addplot[TextMuted, dashed, line width=0.9pt, dash pattern=on 4pt off 2.5pt]
    coordinates {(669.6,0.050237) (19444.4,0.044604)};
  \addlegendentry{Fit: $L = 0.063\,D^{-0.035}$}
  \addplot[only marks, mark=*, mark size=3.2pt, mark options={fill=ScaleA, draw=white, line width=0.6pt}] coordinates {(16151.00,0.045031)};
  \node[font=\sffamily\scriptsize, text=TextBody, inner sep=1pt, anchor=south west] at (axis cs:16151.00,0.045031) [xshift=3pt, yshift=2pt] {100\%};
  \addplot[only marks, mark=*, mark size=3.2pt, mark options={fill=ScaleB, draw=white, line width=0.6pt}] coordinates {(8075.50,0.046030)};
  \node[font=\sffamily\scriptsize, text=TextBody, inner sep=1pt, anchor=south west] at (axis cs:8075.50,0.046030) [xshift=3pt, yshift=2pt] {50\%};
  \addplot[only marks, mark=*, mark size=3.2pt, mark options={fill=ScaleC, draw=white, line width=0.6pt}] coordinates {(4037.75,0.047008)};
  \node[font=\sffamily\scriptsize, text=TextBody, inner sep=1pt, anchor=north east] at (axis cs:4037.75,0.047008) [xshift=-3pt, yshift=-2pt] {25\%};
  \addplot[only marks, mark=*, mark size=3.2pt, mark options={fill=ScaleD, draw=white, line width=0.6pt}] coordinates {(2018.88,0.047965)};
  \node[font=\sffamily\scriptsize, text=TextBody, inner sep=1pt, anchor=north east] at (axis cs:2018.88,0.047965) [xshift=-3pt, yshift=-2pt] {12.5\%};
  \addplot[only marks, mark=*, mark size=3.2pt, mark options={fill=ScaleE, draw=white, line width=0.6pt}] coordinates {(1009.44,0.049855)};
  \node[font=\sffamily\scriptsize, text=TextBody, inner sep=1pt, anchor=south west] at (axis cs:1009.44,0.049855) [xshift=3pt, yshift=2pt] {6.25\%};
  \end{axis}
\end{tikzpicture}
  \caption{Post-training action loss on held-out demonstrations against the
  amount of pre-training data, on logarithmic axes. Each point is the
  16-set average in the right panel of Figure~\ref{fig:scaling} over its
  last 1,000 updates; the dashed line is a least-squares power-law fit
  ($R^2 = \PowerLawRsq$).}
  \label{fig:powerlaw}
\end{figure}
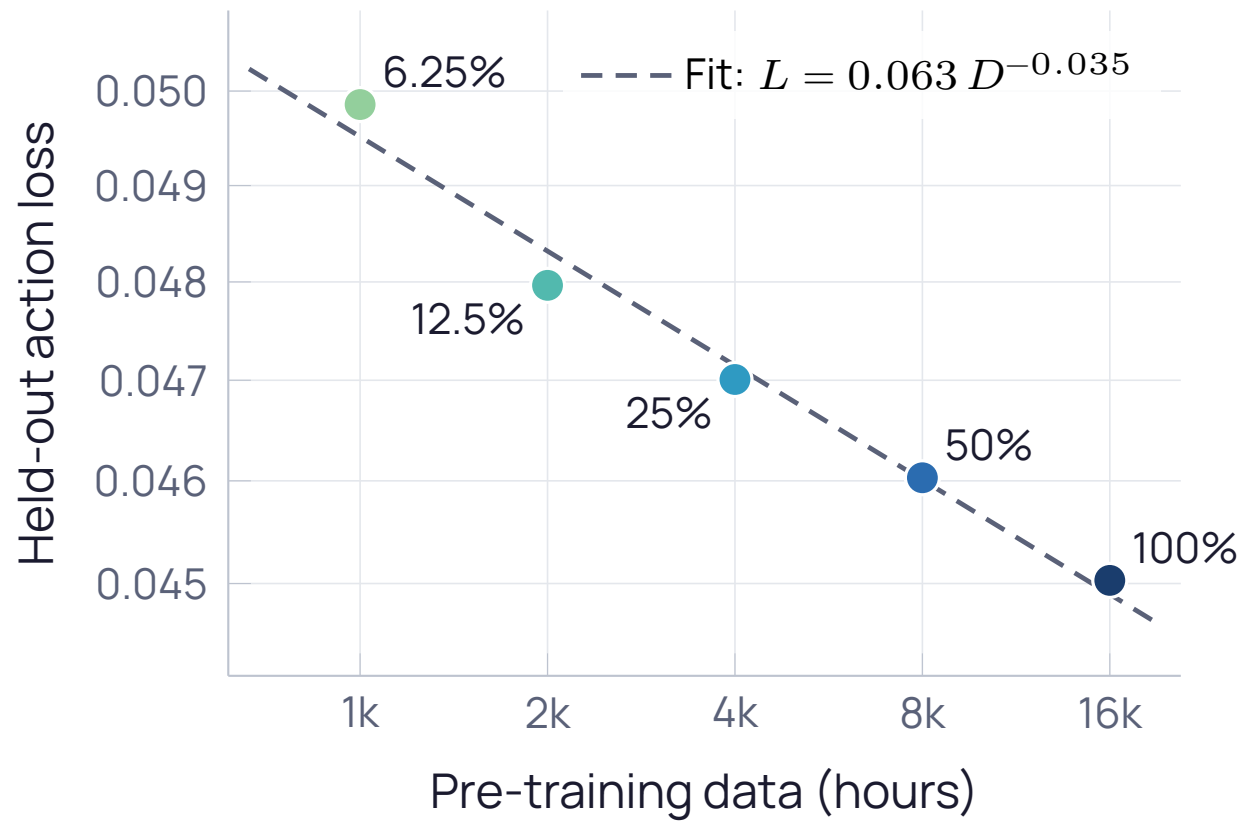

The advantage of more pre-training data survives post-training: given the
same task demonstrations, the models pre-trained on less data do not catch
up. From each run we take the
checkpoint at 100,000 steps and post-train it on the same 16 task sets,
whose demonstrations none of the pre-training runs sampled, with the
pre-training objective and identical settings. Every 50
updates we measure the action loss on held-out demonstrations of each set,
with the subtask instruction as prompt (Figure~\ref{fig:scaling}, right).
Averaged over the sets, the model
pre-trained on more data has the lower loss at every evaluation shown, and
after the first thousand updates the gap between the shares no longer
narrows. Set by set
(Figure~\ref{fig:posttrain-scaling-tasks} in the appendix), the smallest
share has the highest loss on all 16.

The relation is regular enough to fit. Averaged over the last 1,000 updates
in the right panel of Figure~\ref{fig:scaling}, the post-training loss of the five
runs follows a power law in the
amount of pre-training data $D$, $L = \PowerLawA\,D^{-\PowerLawAlpha}$ with
$D$ in hours ($R^2 = \PowerLawRsq$; Figure~\ref{fig:powerlaw}): from about
1,000 to 16,000 hours, each doubling of the pre-training data lowers the loss
after post-training by about \PowerLawDoubling\%.

\Needspace{6\baselineskip}
\section{Post-training}
\label{sec:posttraining}

Post-training makes the pre-trained model better at the tasks it will
perform, and it is where \mbox{A-0} acquires what long tasks demand: knowing
how far the task has progressed, and remembering what the arms have just
done. It trains on handheld demonstrations of those tasks and adds two
modules, a stage head and a proprioceptive memory
(Figure~\ref{fig:model}).

\subsection{Stage prediction}

A long task is a sequence of subtasks, and the two wrist views at a given
moment are often consistent with more than one of them. A policy that is
given only the task instruction has to resolve this ambiguity inside its
action distribution. \mbox{A-0} resolves it explicitly: each task is divided
into stages, each a run of consecutive subtasks (Table~\ref{tab:levels}), and
the model predicts the current stage before it acts.

\begin{table}[H]
  \centering
  \caption{The parts-kitting task at three levels of language. The task
  instruction is all the robot is given; the stage is predicted by the model;
  the subtask instructions label the demonstrations.}
  \label{tab:levels}
  \vspace{0.5em}
  \small
  \begin{tabular}{@{}p{0.07\textwidth} >{\raggedright\arraybackslash}p{0.33\textwidth}
      >{\raggedright\arraybackslash}p{0.52\textwidth}@{}}
    \toprule
    \textbf{Task} & \multicolumn{2}{p{0.87\textwidth}@{}}{Place each specified YUBI part into its corresponding numbered cell.} \\
    \midrule
    & \textbf{Stage} & \textbf{Subtasks} \\
    \cmidrule(l){2-3}
    & Place the fingers, flaps and gears \newline in cells 1, 2 and 3. &
      Pick the finger part and place it into cell 1. \newline
      Pick the flap part and place it into cell 2. \newline
      Pick the gear part and place it into cell 3. \\[4pt]
    & Place the shafts, gear boxes and \newline Quest holder in cells 4, 5 and 6. &
      Pick the shaft part and place it into cell 4. \newline
      Pick the gear box part and place it into cell 5. \newline
      Pick the Quest holder part and place it into cell 6. \\[4pt]
    & Place the camera brackets and grip \newline in cells 7 and 8. &
      Pick the camera bracket part and place it into cell 7. \newline
      Pick the grip part and place it into cell 8. \\
    \bottomrule
  \end{tabular}
\end{table}

A stage head reads the backbone's representation of the observation, and the
predicted stage is passed to the action expert as language, in the form in
which pre-training taught it to follow subtask instructions. The head is
trained with a cross-entropy loss $\Lc_{\text{stage}}$, whose targets come
from the subtask labels of the demonstrations at no extra cost; on the robot its
prediction is smoothed over time, as Section~\ref{sec:ema} describes.

\subsection{Proprioceptive memory}

A single observation does not say what the arms have just done. \mbox{A-0} adds a
memory of recent motion: the last ten proprioceptive states, each with the
time elapsed since it was measured. Four learned queries, conditioned on the
instruction and the current state, attend over this history and compress it
into four tokens that join the prompt.

Robots query a policy at different rates, and a memory indexed by step would
mean something different on each of them. The history is therefore sampled
in training at rates from 5 to 15\,Hz, and the true age of every state is
encoded, so that one policy serves any query rate in that range. The history
is also withheld a quarter of the time, so the policy uses the memory when it
is present and remains able to act without it. At run time the memory holds
measured states only.

\subsection{Visual evidence budget in post-training}

Post-training keeps the visual budget gate and uses it in two places. As in
pre-training, the gate scales the image tokens that enter the backbone; for
post-training the budget is $\rho = 0.9$, the temperature $\tau = 0.805$, and
the floor is annealed from 0.1 to 0.02. In addition, the gate acts on attention. Its keep
probabilities, with gradients stopped and projected so that they respect the
budget ($\tilde{m}_p$), become a prior $b_p = \log\bigl(e^{-\kappa} + (1 - e^{-\kappa})\,\tilde{m}_p\bigr)$, between $-\kappa$ and 0, that is added to
the attention of action tokens over image tokens,
\begin{equation}
  A = \operatorname{softmax}\!\Bigl(\frac{QK^{\top}}{\sqrt{d}} + \beta\, b\Bigr),
  \label{eq:prior}
\end{equation}
with $\kappa = 4$ and $\beta$ raised from 0 to 1 over the first 500 steps. The
gate is thus trained through the embeddings, by the action loss, and applied
a second time where the action expert reads the image. Both are active at
run time.

\subsection{Objective and recipe}

The post-training objective takes the flow-matching action loss
$\Lc_{\text{act}}$~\cite{pi0}, computed through the gate, and adds the stage
loss to the budget and subtask terms of pre-training:
\begin{equation}
  \Lc_{\text{post}} = \Lc_{\text{act}}
  + \lambda_{\text{b}}\, \Lc_{\text{budget}}
  + \lambda_{\text{c}}\, \Lc_{\text{stage}}
  + \lambda_{\text{s}}\, \Lc_{\text{sub}}.
  \label{eq:post}
\end{equation}
Table~\ref{tab:posttrain} gives the recipe, loss weights included. The
same recipe serves every task we report, and a run takes under four hours on
a single node of eight H200 GPUs.

\begin{table}[H]
  \centering
  \caption{Post-training recipe.}
  \label{tab:posttrain}
  \vspace{0.5em}
  \small
  \begin{tabular}{@{}l l@{}}
    \toprule
    Initialisation     & pre-trained weights; stage head and memory from scratch \\
    Action chunk       & 20 anchored actions at 20\,Hz \\
    Prompt             & task instruction \\
    Visual evidence budget & $\rho = 0.9$, $\gamma$: $0.1 \rightarrow 0.02$, $\tau = 0.805$, $\lambda_{\text{b}} = 0.1$ \\
    Loss weights       & stage $\lambda_{\text{c}} = 0.1$, subtask $\lambda_{\text{s}} = 0.01$ \\
    Proprioceptive memory & ten past states, four memory tokens \\
    Updates / batch size & 20,000 / 64 \\
    Optimiser          & AdamW, new optimiser state, gradient clipping at 1 \\
    Learning rate      & $5\times10^{-6}$, constant after 500 warm-up steps \\
    Weights deployed   & exponential moving average, decay 0.99 \\
    \bottomrule
  \end{tabular}
\end{table}

\Needspace{6\baselineskip}
\section{Inference Harness}
\label{sec:inference}

On the robot, a policy is queried a few times a second and has to produce
motion that is continuous, consistent with what it did a moment ago, and
right about where the task stands. The inference harness turns the model's
predictions into such motion with two main components, an exponential moving
average on the predicted stage and a streaming decoder for actions.

\subsection{Stage smoothing}
\label{sec:ema}

The stage is predicted afresh at every query from two wrist views that show
only part of the scene, and one wrong prediction would hand the action expert
the instruction for a different step. The task itself, meanwhile, passes from
one stage to the next gradually. The harness therefore conditions the action
expert on the most probable stage under an exponential moving average of the
predicted probabilities, the stage EMA. An isolated misprediction barely
moves it, a real transition is followed as it unfolds, and the filter needs no
retuning when the query rate changes.

\subsection{Non-uniform streaming decoding}
\label{sec:stream}

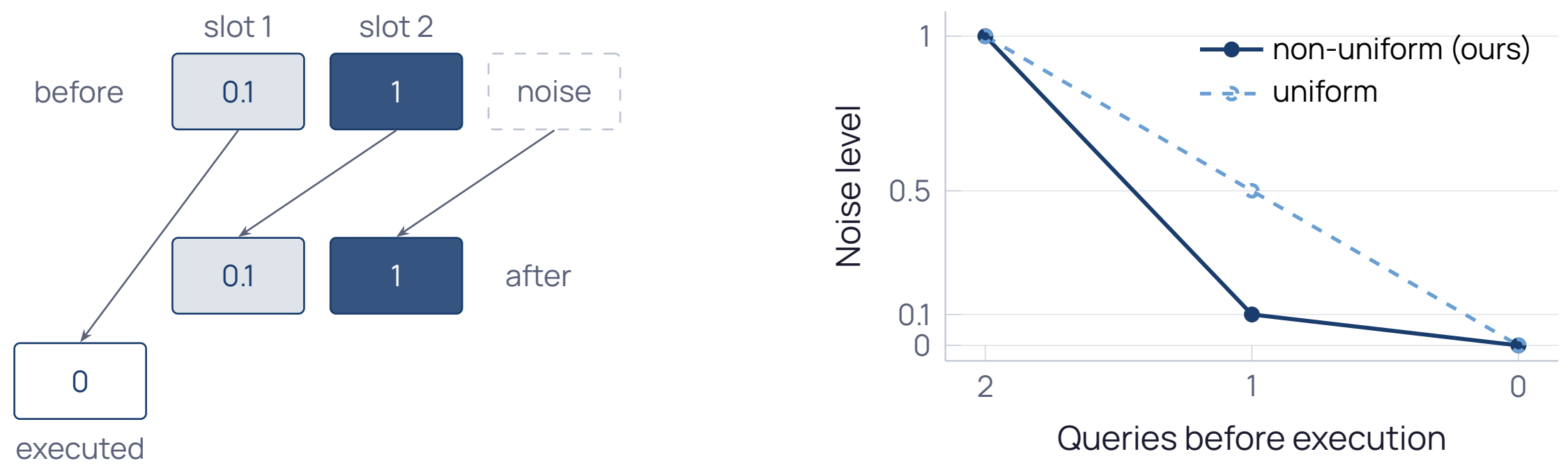
\begin{figure}[H]
  \centering
  \begin{minipage}[c]{0.50\textwidth}
    \centering
    \begin{tikzpicture}[
        slot/.style={rounded corners=1.5pt, minimum width=1.25cm,
          minimum height=0.72cm, font=\sffamily\scriptsize, inner sep=1pt,
          draw=Primary, line width=0.5pt},
        lab/.style={font=\sffamily\scriptsize, text=TextMuted},
        arr/.style={-{Stealth[length=3.5pt]}, draw=TextMuted, line width=0.5pt}]
      \node[lab, anchor=east] at (0.55, 0) {before};
      \node[slot, fill=Primary!14, text=Primary] (a1) at (1.5, 0) {0.1};
      \node[slot, fill=Primary!88, text=white]   (a2) at (3.0, 0) {1};
      \node[slot, draw=RuleGray, dashed, text=TextMuted] (new) at (4.5, 0) {noise};
      \node[lab, anchor=west] at (3.9,-1.75) {after};
      \node[slot, fill=white, text=Primary] (b0) at (0.0,-2.75) {0};
      \node[slot, fill=Primary!14, text=Primary] (b1) at (1.5,-1.75) {0.1};
      \node[slot, fill=Primary!88, text=white]   (b2) at (3.0,-1.75) {1};
      \draw[arr] (a1.south) -- (b0.north);
      \draw[arr] (a2.south) -- (b1.north);
      \draw[arr] (new.south) -- (b2.north);
      \node[lab, anchor=north] at (0.0,-3.15) {executed};
      \node[lab] at (1.5, 0.62) {slot 1};
      \node[lab] at (3.0, 0.62) {slot 2};
    \end{tikzpicture}
  \end{minipage}%
  \hfill
  \begin{minipage}[c]{0.47\textwidth}
    \centering
    \begin{tikzpicture}
      \begin{axis}[
          width=7.4cm, height=4.9cm,
          xlabel={Queries before execution},
          ylabel={Noise level},
          xmin=-0.15, xmax=2.15, ymin=-0.05, ymax=1.08,
          xtick={0,1,2}, xticklabels={2,1,0},
          ytick={0,0.1,0.5,1},
          tick label style={font=\sffamily\scriptsize, text=TextMuted},
          label style={font=\sffamily\footnotesize, text=TextBody},
          axis line style={RuleGray}, tick style={RuleGray},
          ymajorgrids, grid style={RuleGray!45, line width=0.3pt},
          axis x line*=bottom, axis y line*=left,
          legend style={font=\sffamily\scriptsize, draw=none, fill=none,
            at={(0.99,0.98)}, anchor=north east, legend cell align=left},
        ]
        \addplot[Primary, line width=1.1pt, mark=*, mark size=1.6pt]
          coordinates {(0,1) (1,0.1) (2,0)};
        \addlegendentry{non-uniform (ours)}
        \addplot[AccentMuted, line width=1.0pt, dashed, mark=o, mark size=1.6pt]
          coordinates {(0,1) (1,0.5) (2,0)};
        \addlegendentry{uniform}
      \end{axis}
    \end{tikzpicture}
  \end{minipage}
  \caption{Streaming decoding with two slots. Left: the buffer before and
  after one query. Right: the noise level of one chunk on its way through the
  buffer, with a uniform schedule and with the non-uniform schedule used in
  \mbox{A-0}.}
  \label{fig:stream}
\end{figure}

Standard flow-matching inference generates each action chunk from pure noise
with ten evaluations of the action expert, and each chunk independently of
the last. Streaming decoding, which FlashVLA~\cite{flashvla} brings to
vision-language-action models, instead
holds a buffer of $N$ chunks at staggered noise levels
$0 = \ell_0 < \ell_1 < \dots < \ell_N = 1$, with attention causal from one
chunk to the next. At each query the action expert denoises the whole buffer
on the newest observation: the chunk in slot $k$ moves from $\ell_k$ to
$\ell_{k-1}$, the first comes out clean and is executed, and fresh noise
enters at the back (Figure~\ref{fig:stream}, left). Every chunk is thus
refined over $N$ successive observations, in the context of the chunks
before it.

With evenly spaced levels, as in FlashVLA, the size of every update is tied to
the number of slots. With few slots the update just before execution is large,
and on the robot this shows as seams between consecutive chunks; with many,
each chunk is settled on observations taken long before it runs, and the
policy is slow to respond when the scene changes.

We propose non-uniform streaming decoding, which makes the size of
an update depend on how far the chunk is from execution
(Figure~\ref{fig:stream}, right). \mbox{A-0} uses two slots, with levels
$(0.1,\, 1)$: a chunk entering the buffer takes the large step from 1 to 0.1
while the chunk ahead is nearly clean, and just before it runs it receives
only a small correction from 0.1 to 0 on the newest observation. The motion
stays continuous from one chunk to the next even with so few slots, and the
policy still responds promptly when the scene changes
(Section~\ref{sec:perturb}). Since every action is anchored at the pose of
the query, the harness keeps each buffered chunk consistent with the newest
measured pose and with the current stage.

In training, the noise level of the chunk in slot $k$ is drawn from that
slot's own interval,
\begin{equation}
  t_k = \ell_{k-1} + (\ell_k - \ell_{k-1})\, u, \qquad u \sim \mathrm{Beta}(1.5,\, 1),
  \label{eq:slot}
\end{equation}
and training covers both steady-state buffers and cold starts. The decoder
adds no parameters. The noise level of each chunk enters through the time
conditioning the action expert already has, and a streaming policy is
obtained from the same pre-trained model with the same post-training recipe.

Streaming also makes each query cheaper. The observation is encoded once,
and in steady state the action expert is evaluated twice, against
ten times for standard decoding; a query takes about 60\,ms on an H200 GPU.

\section{Related Work}
\label{sec:related}

\textbf{Handheld data collection.} UMI~\cite{umi} showed that a handheld
gripper with a wrist camera can record demonstrations that transfer directly
to robots, with no robot present when the data is collected.
Dobb-E~\cite{dobbe} and Robot Utility Models~\cite{rum} pursued the same goal
in homes with a handheld tool, the latter deploying in unseen environments
with no further training, and Lin et al.~\cite{scalinglaws} found that
generalisation to new environments and objects follows a power law in the
number of environments and objects recorded, which is what handheld
collection provides most readily. YUBI~\cite{yubi} is our gripper
of this kind, finger-driven and designed for bimanual work and for
collection at scale. Handheld data has since reached the scale of
foundation-model pre-training. RDT2~\cite{rdt2} reports zero-shot
generalisation to unseen objects, scenes, instructions and robots, and
\mbox{Xiaomi-Robotics-1}~\cite{xiaomi} finds that the gains from data and model
scale in pre-training carry over to real-robot performance after
post-training. Two industrial systems start from the same premise:
GEN-1~\cite{gen1} pre-trains on data recorded with wearable devices, with no
robot data, and adapts to each task with a small amount of robot data, and
OM-1~\cite{om1} reports learning from human data alone, recorded
with an instrumented wearable hand. HiFi-UMI~\cite{hifiumi} approaches the
question from the data: it raises the fidelity of the recordings, releases
2,000 hours of them, and reports that a policy post-trained on them alone
matches one trained on teleoperation recorded in the evaluation scene itself.
RealOmni-Open~\cite{realomni} reports more than 13,000 hours of open
household recordings made with a handheld gripper. \mbox{A-0} builds on this
line and carries it to long bimanual tasks specified by a single
instruction, seen through wrist cameras alone and carried out at a site
outside the data.

\textbf{Vision-language-action models.} RT-2~\cite{rt2} named the family, and
the open generalist policies Octo~\cite{octo} and OpenVLA~\cite{openvla}, like
it, learn from data recorded on robots, pooled across laboratories in Open
X-Embodiment~\cite{oxe} or collected in the wild in DROID~\cite{droid}, so the
data grows with the number of robots.
\piz{}~\cite{pi0} and \pio{}~\cite{pi05} pair a pre-trained vision-language
model with an action expert trained by flow matching; GR00T~N1~\cite{groot}
and Gemini Robotics~\cite{gemini} are further members of the family.
GEN-0~\cite{gen0} reports that such models improve predictably with the
scale of real-world interaction data, and $\pi^{*}_{0.6}$~\cite{pistar}
improves such a model further by reinforcement learning from on-robot
experience and corrections. \mbox{A-0} adopts the \pio{} architecture and its public weights as
a starting point, and collects neither robot demonstrations nor on-robot
experience.

\textbf{Language and long tasks.} Long tasks are commonly handled by placing
language between the instruction and the action. RT-H~\cite{rth} predicts a
language motion before each action, Hi~Robot~\cite{hirobot} has a high-level
model choose the next step for a low-level policy, and \pio{} predicts a
subtask in free text. S2~\cite{s2} trains the policy itself on language that
specifies the current step. \mbox{A-0} follows S2 in pre-training and, in
post-training, reduces the high-level decision to a classification over
stages, in the manner of the stage tracking with which Larchenko et
al.~\cite{behavior} took first place in the 2025 BEHAVIOR Challenge; the
subtask labels of its data make this possible.
History is the other source of the context a long task needs, and imitation
policies tend to under-use it~\cite{pasttoken}; \mbox{A-0} keeps its history short
and proprioceptive.

\textbf{Selective perception.} S2~\cite{s2} trains a policy under a visual
evidence budget, learned from the control objective with no visual
annotation, and reports robustness gains from it that add to those of more
specific language. BYOVLA~\cite{byovla} seeks robustness to visual distractors at
run time instead, editing task-irrelevant regions of the image with external
tools. \mbox{A-0} applies the budget to its corpus and carries the gate through
post-training into deployment.

\textbf{Real-time action generation.} Action chunking with transformers
(ACT)~\cite{act} and diffusion policies~\cite{diffusionpolicy} predict a
sequence of actions at once, which leaves consecutive sequences to be
reconciled when they are executed; ACT averages the chunks that overlap. Real-time chunking~\cite{rtc} does so at
inference time, generating each chunk consistently with the unexecuted part
of the last. A different route denoises a buffer whose entries sit at
different noise levels: Diffusion Forcing~\cite{dforcing} trains sequence
models with an independent noise level per token, and Streaming Diffusion
Policy~\cite{sdp} keeps a buffer of robot actions in which later actions are
noisier, so that each new observation needs only a few denoising steps.
FlashVLA~\cite{flashvla} brings this to vision-language-action models by
streaming chunks through a buffer of staggered noise levels, at 30\,Hz or
more on a single GPU. \mbox{A-0} builds on
FlashVLA with the non-uniform schedule we propose here, and adapts it to
anchored actions and to stage-conditioned policies.

\section{Conclusion}
\label{sec:conclusion}

\mbox{A-0} learns bimanual manipulation from handheld demonstrations, as one
system that runs from data processing to the inference harness on the robot.
With no robot data collected, it carries out long, precise tasks from a
single instruction, through two wrist cameras, at a site outside its data,
and it runs on robots that carry the same grippers. The corpus is still growing, and
\mbox{A-0} is the first system we have built on it.

\section*{Contributions}

\textbf{Leadership.} Yueh-Hua (Kris) Wu led the research, and he and Kei Ota
led the project.

\textbf{System.} Yueh-Hua (Kris) Wu conceived \mbox{A-0}, and designed and
implemented its data processing, pre-training, post-training and inference
harness.

\textbf{YUBI.} Takehiko Ohkawa led the YUBI project, which Kei Ota and
Tatsuya Matsushima supervised. Yuki Noguchi co-designed the gripper, and
Makoto Sugiura led its refinement and manufacture. Naoaki Kanazawa, Petr
Khrapchenkov, Takuya Okubo and Yuki Noguchi developed the data-collection
software, and Makoto Sugiura, Takuya Okubo and Yuki Noguchi integrated the
gripper with robot platforms.

\textbf{Data.} Takehiko Ohkawa managed data collection and task
definition together with Kengo Ikeuchi, Kensuke Chiyoma, Masatoshi Tateno and
Yuma Shin. Masafumi Takahashi
and Petr Khrapchenkov built the data infrastructure, and Masafumi Takahashi
also curated the task datasets. Masatoshi Tateno generated the language
annotations.

\textbf{Evaluation.} Hanlong Li, Hayata Yuasa, Jason Wang, Masatoshi Tateno
and Yueh-Hua (Kris) Wu carried out the evaluations on the bimanual Franka,
where Masatoshi Tateno also evaluated the baseline models. Hayata Yuasa,
Satoshi Otsubo and Zixuan Zhong carried out those on TX-G2, Nicola Scianca
those on OpenArm, and Yuki Noguchi those on UR.

\section*{Acknowledgments}

This paper is based on results obtained from a project, JPNP25015,
commissioned by the New Energy and Industrial Technology Development
Organization (NEDO).

\clearpage
\appendix
\section{Post-training Loss by Task Set}
\label{sec:appendix}

\begin{figure}[!htb]
  \centering
  \input{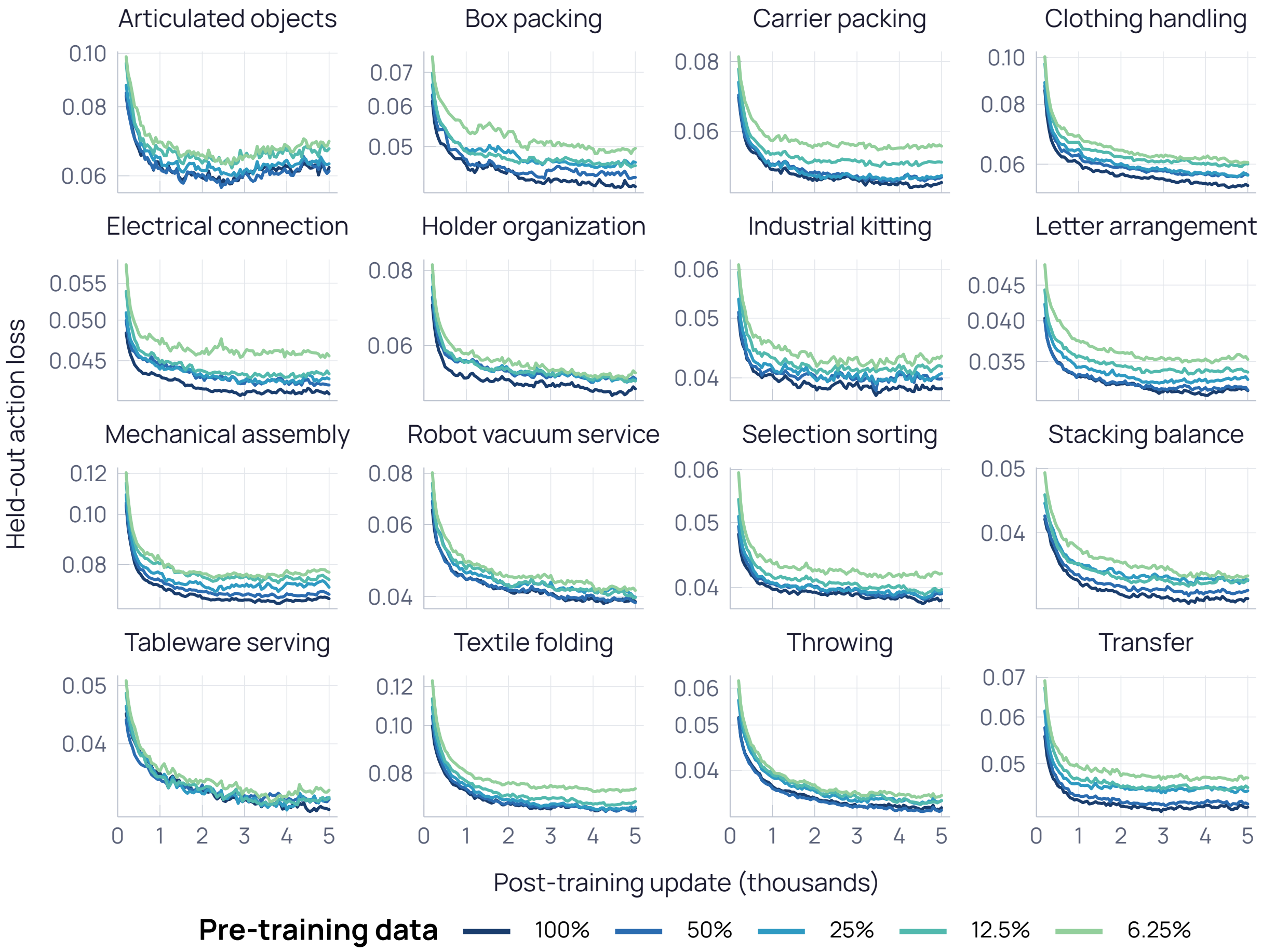}
  \caption{The loss in the right panel of Figure~\ref{fig:scaling} for
  each of the 16 task sets.}
  \label{fig:posttrain-scaling-tasks}
\end{figure}

Figure~\ref{fig:posttrain-scaling-tasks} gives the post-training loss in
the right panel of Figure~\ref{fig:scaling} for each of the 16 task sets
(Section~\ref{sec:scaling}).

\end{document}